\documentclass[10pt]{article}

\usepackage[preprint]{tmlr}

\usepackage{amsmath,amsfonts,bm}

\def\eqref#1{equation~\ref{#1}}

\def\1{\bm{1}}

\DeclareMathAlphabet{\mathsfit}{\encodingdefault}{\sfdefault}{m}{sl}
\SetMathAlphabet{\mathsfit}{bold}{\encodingdefault}{\sfdefault}{bx}{n}

\usepackage{amsmath,amssymb}
\usepackage{booktabs}
\usepackage{array}
\usepackage{multirow}
\usepackage{tabularx}
\usepackage{longtable}
\usepackage[hypertexnames=false,hidelinks]{hyperref}
\usepackage{enumitem}
\usepackage{graphicx}
\usepackage{float}


\usepackage{threeparttable}
\usepackage{makecell}
\usepackage{xcolor}

\usepackage{tikz}
\usetikzlibrary{arrows.meta,positioning,fit}

\begin{document}

\raggedbottom

\setlength{\abovedisplayskip}{6pt}
\setlength{\belowdisplayskip}{6pt}
\setlength{\abovedisplayshortskip}{4pt}
\setlength{\belowdisplayshortskip}{4pt}


\title{ETH-TraceBench: A Large-Scale Event-Stream Benchmark for Ethereum DeFi under Temporal, Protocol, and Contract Shift}

\author{
\begin{center}
{\Large\textbf{Kemal Kirtac}$^{a,b}$ \qquad \textbf{Carsten Maple}$^{c}$}\\[0.9em]
{\normalfont\normalsize
$^{a}$Department of Computer Science, University College London\\
66--72 Gower Street, London WC1E 6EA, United Kingdom\\[0.5em]
$^{b}$Department of Computer Science, University of Warwick\\
Coventry CV4 7AL, United Kingdom\\[0.5em]
$^{c}$WMG, University of Warwick\\
Coventry CV4 7AL, United Kingdom\\[0.8em]
{\small\itshape Under review at Transactions on Machine Learning Research (TMLR).}
}
\end{center}
}

\maketitle

\begin{abstract}
{\color{black}
Ethereum decentralized finance (DeFi) provides a public, time-stamped record of transaction-level event streams, but the same public symbols that make the data observable can create strong machine-learning shortcuts. We introduce \textsc{ETH-TraceBench}, a benchmark for evaluating Ethereum DeFi representations under naturally occurring temporal, protocol, pool/infrastructure, and symbolic shift.

The raw event universe covers January 2021--December 2025 and contains 1.35 billion transactions with logs and 5.01 billion raw log rows. The positive-label source population contains 311.87 million DEX transactions and 105,952 liquidation transactions, with 2.33 billion and 2.45 million matched logs, respectively. Model evaluation uses a fixed 911,267-instance supervised sample containing source positives and source-excluded comparison transactions. The supervised protocol trains on 2021--2024, selects models on 2025H1, and evaluates on 2025H2. Simple models are deliberately strong on the aggregate temporal test: on the exact canonical DEX test set, TraceStats-GB reaches 0.953 macro-F1 and TopicEmitterHashMLP reaches 0.959. The benchmark becomes substantially harder under protocol novelty: macro-F1 falls to 0.794, 0.743, and 0.766 for TraceStats-GB, TopicEmitterTrace-SGD, and TopicEmitterHashMLP, respectively, on the strict unseen-protocol subset, while the corresponding strict unseen-pool scores remain 0.927, 0.897, and 0.935. Named transfer is also difficult: Uniswap v4 and Ekubo v1 are absent from supervised training and yield materially lower scores than the full test. Masking experiments show that emitter-contract identity alone does not explain DEX performance, whereas jointly masking emitter and topic identity reduces DEX macro-F1 to 0.916 and liquidation macro-F1 to 0.774 for TopicEmitterTrace-SGD. A standard Transformer evaluated on log-index-ordered events provides no consistent advantage over a deterministic shuffle of the same retained events, showing that high aggregate scores can arise without sophisticated chronological modeling. A natural-prevalence audit estimates 2025H2 DEX prevalence among logged Ethereum transactions at approximately 22.5\% and shows that the canonical DEX sample closely tracks the population by month, project, protocol, and trace complexity. A deterministic 400-transaction source-concordance and event-signature audit finds complete agreement with the task label sources and independently re-queried raw-log counts. \textsc{ETH-TraceBench} therefore uses difficult transfer and controlled-input conditions, rather than a single aggregate score, as the main evaluation target.
}
\end{abstract}

\section*{Abbreviations}
\noindent DeFi: decentralized finance; EVM: Ethereum Virtual Machine; MEM-BL: Masked Event Modeling for Blockchain Logs; AMM: automated market maker; ABI: application binary interface.

{\color{black}
\section{Introduction}
\label{sec:intro}

Ethereum decentralized finance (DeFi) produces a large public record of executable economic activity \citep{Schar2021,Werner2022}. Each transaction is time-stamped and can emit an ordered sequence of receipt logs whose event topics and emitting contracts record parts of the transaction's execution \citep{Wood2014Ethereum}. Receipt logs are distinct from complete EVM call trees or opcode-level execution traces \citep{Wood2014Ethereum,Gai2023BlockGPT}, but they provide a protocol-agnostic event-stream object that can be reconstructed at ledger scale. At ledger scale, this event stream supports a general machine-learning question: whether representations learned from historical structured events remain reliable when the protocols, contracts, pools, and symbolic vocabularies generating those events change.

The same observability also creates an unusually strong shortcut problem. Stable event signatures, emitting-contract identities, generic token-transfer topics, and aggregate trace complexity can be highly predictive even when they do not encode reusable transaction semantics. A model can therefore perform extremely well on a conventional future-period test while remaining brittle to genuinely new protocol mechanisms or symbolic structure. The shortcut problem connects blockchain learning to the broader literature on shortcut learning and evaluation under naturally occurring distribution shift \citep{Geirhos2020Shortcut,Koh2021Wilds,Gulrajani2021DomainBed,Sagawa2020GroupDRO}.

\textsc{ETH-TraceBench} is designed around this shortcut problem. The raw universe covers Ethereum mainnet from January 2021 through December 2025 and contains 1.35 billion transactions with logs and 5.01 billion raw log rows. The task-labelled benchmark contains 311.97 million labelled transaction instances and 2.33 billion matched raw logs. DEX swap/trade classification is the main large-scale task; liquidation detection is a rare-event stress test. The labels are derived from curated on-chain activity tables, while the model inputs are constructed from raw-log event streams rather than decoded project or protocol fields.

Five evaluation settings organize the analysis. \emph{Temporal evaluation} trains on 2021--2024, selects models on 2025H1, and tests on 2025H2. \emph{Protocol transfer} evaluates transactions from project/protocol combinations absent from supervised training. \emph{Pool holdout} evaluates transactions associated with pool or liquidity-infrastructure identifiers absent from training. \emph{Symbolic novelty} separates test transactions containing unseen event topics or unseen emitter--topic pairs. \emph{Masking} removes emitter-contract identity, event-topic identity, or both from the model input. Here and throughout the paper, an ``emitter address'' means the smart-contract address in \texttt{ethereum.raw.logs.address}; it is not a trader, investor, borrower, liquidator, or transaction-initiator wallet.

Aggregate temporal performance and hard-transfer performance diverge sharply. On the exact canonical DEX test set, simple models remain strong: TraceStats-GB reaches 0.953 macro-F1, TopicEmitterTrace-SGD reaches 0.924, and TopicEmitterHashMLP reaches 0.959. However, strict unseen-protocol macro-F1 falls to 0.794, 0.743, and 0.766, respectively. Strict unseen-pool performance remains much higher at 0.927, 0.897, and 0.935. Thus protocol novelty is substantially harder than local pool novelty even when the aggregate temporal score appears close to saturation. Named new-protocol evaluation shows the same pattern: Uniswap v4 and Ekubo v1, both absent from training, are materially harder than the full test.

Shortcut-controlled experiments provide a second diagnostic. For TopicEmitterTrace-SGD, masking emitter-contract identity alone does not reduce DEX macro-F1, while joint emitter+topic masking lowers DEX macro-F1 to 0.916 and liquidation macro-F1 to 0.774. A standard Transformer is evaluated over chronologically ordered raw-log events. Comparing the true order with a deterministic shuffle of the same retained events shows no consistent ordering advantage: DEX macro-F1 is 0.943 ordered versus 0.953 shuffled, and liquidation macro-F1 is 0.947 versus 0.948. The comparison indicates that the high scores primarily reflect event composition and symbolic structure rather than sophisticated chronological reasoning.

Benchmark validity is assessed along three additional dimensions. A natural-prevalence audit estimates that DEX-labelled transactions comprise approximately 22.5\% of Ethereum transactions with logs in 2025H2 and shows that the canonical positive sample closely tracks the full DEX population by month, project, protocol, and trace complexity. Infrastructure-adjacent hard negatives provide a second check: 76.4\% of canonical DEX negatives and 76.1\% of liquidation negatives share at least one exact emitter--topic pair with a training positive, and these negatives are substantially more confusable for several baselines. A deterministic 400-transaction source-concordance audit---including ordinary and hard negatives---finds 400/400 agreement with the source-table label rule, 400/400 raw-log presence, and 400/400 exact agreement between canonical and independently re-queried log counts. An event-signature consistency check finds a positive task signature in all 200 sampled positives and none of the 200 sampled negatives. The source-concordance and semantic-consistency checks are automated and are not independent human annotation.

ETH-TraceBench contributes a ledger-scale, transaction-level raw-log benchmark with fixed chronological partitions and reproducible task labels. Protocol transfer, pool holdout, symbolic novelty, and masking are treated as empirical evaluation conditions rather than future leaderboard requirements. The empirical package adds same-instance baseline comparisons, a standard sequence model, paired uncertainty estimates, hard-negative stress tests, and prevalence/representativeness diagnostics that expose when aggregate performance is misleading. An auditable release protocol ties numerical claims to transaction-level prediction artifacts, split manifests, and regeneration code. The benchmark therefore tests whether representation learning remains reliable under naturally occurring changes in the infrastructure and symbols that generate DeFi event streams.

Figure~\ref{fig:eth-tracebench-pipeline} summarizes how the benchmark separates construction, labels, temporal partitions, shift manifests, and evaluation artifacts. Separating label-source metadata from predictive inputs is important because project and protocol fields are used for supervision and diagnostics but are not supplied to the reported models.

\begin{figure}[!htbp]
\centering
\includegraphics[width=0.92\textwidth]{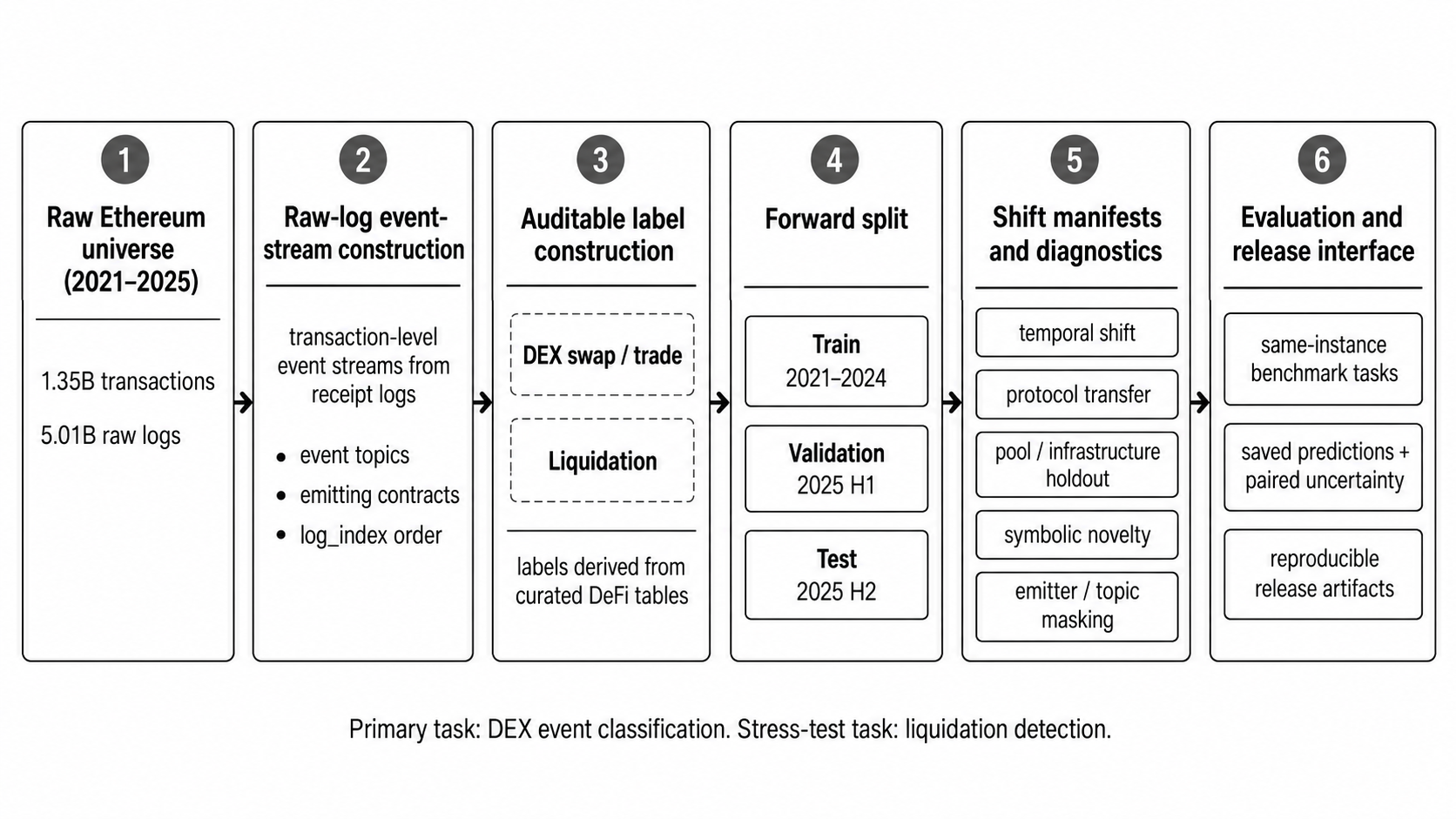}
\caption{ETH-TraceBench benchmark construction and evaluation pipeline. The raw Ethereum universe is converted into transaction-level event streams from receipt logs; DEX and liquidation labels are derived from curated DeFi tables; chronological partitions are fixed before model evaluation; and temporal shift, protocol transfer, pool/infrastructure holdout, symbolic novelty, and emitter/topic masking are evaluated from reproducible manifests and prediction artifacts.}
\label{fig:eth-tracebench-pipeline}
\end{figure}

}

{\color{black}
\section{Related Work and Benchmark Positioning}
\label{sec:related}

\subsection{Real-world shift, shortcut learning, and benchmark design}
WILDS makes naturally occurring domain and subpopulation shift a first-class evaluation object \citep{Koh2021Wilds}; DomainBed emphasizes standardized datasets, model selection, and evaluation protocols for domain generalization \citep{Gulrajani2021DomainBed}; and GroupDRO motivates worst-group reporting when aggregate performance hides failures on shifted groups \citep{Sagawa2020GroupDRO}. Shortcut learning provides the complementary warning that high predictive accuracy can arise from features that correlate with labels without supporting robust transfer \citep{Geirhos2020Shortcut}. EarthquakeNPP is a useful benchmark-design example because leakage correction, omitted difficult cases, and credible baselines are treated as part of the scientific contribution rather than as post-hoc engineering \citep{Stockman2026EarthquakeNPP}. ETH-TraceBench applies these principles to public blockchain event streams by separating temporal shift, protocol novelty, infrastructure novelty, symbolic novelty, and input masking instead of collapsing them into one future-period score.

The benchmark also follows dataset-documentation work that treats motivation, provenance, preprocessing, uses, limits, and maintenance as part of the scientific object \citep{Gebru2021Datasheets,Bender2018DataStatements}. Datasheets for Datasets and Data Statements motivate structured documentation; Croissant provides machine-readable metadata for ML-ready datasets \citep{Akhtar2024Croissant}; and recent work on benchmark repositories emphasizes persistent versions, checksums, discoverability, and stewardship \citep{Longjohn2024Repositories,Wu2024DatasetManagement}. The release design therefore uses versioned manifests, split hashes, extraction provenance, and prediction-artifact requirements; Appendix~\ref{app:reproducibility} gives the reproducibility checklist and Appendix~\ref{app:repo-structure} gives the release-package structure.

\subsection{Ethereum transaction semantics and execution-aware analytics}
Blockchain machine learning spans address and transaction graphs, smart-contract state, source code and bytecode, decoded protocol records, receipt logs, and internal execution traces; these data objects should not be treated as interchangeable \citep{Azad2026BlockchainML}. TXSPECTOR derives transaction-level logic relations for attack analysis \citep{Zhang2020TXSPECTOR}; BlockGPT learns a domain-specific language model over EVM execution-trace trees \citep{Gai2023BlockGPT}; TxT studies reproducible Ethereum execution paths \citep{Ivanov2023TxT}; MoTS learns generic transaction-semantic representations from recurring interaction motifs \citep{Wu2023MoTS}; DeFiRanger recovers high-level DeFi semantics from transaction cash-flow structure to detect price-manipulation attacks \citep{Wu2024DeFiRanger}; and DeFiGuard constructs transaction cash-flow graphs and applies graph neural networks to price-manipulation detection \citep{Wang2024DeFiGuard}. Graph representation learning has also been used to identify functionally similar DeFi service building blocks from contracts co-used within transactions \citep{Luo2024DeFiServices}.

Collectively, prior execution-aware systems show that transaction-internal structure can support learned transaction representations and downstream detection \citep{Gai2023BlockGPT,Wu2023MoTS,Wu2024DeFiRanger,Wang2024DeFiGuard,Luo2024DeFiServices}. ETH-TraceBench contributes a complementary benchmark object: a fixed, ledger-scale corpus of transaction-level \emph{receipt-log event streams} with explicit future-protocol, future-pool, symbolic-novelty, masking, and same-instance uncertainty evaluations. Ethereum receipt logs preserve emitted topics, emitting-contract addresses, and log order as part of the transaction receipt/log series \citep{Wood2014Ethereum}; they are not a complete EVM call tree. We therefore reserve ``execution trace'' for call-level or opcode-level data. Internal calls, reconstructed token flows, and call depth can be added as derived channels, but they are not conflated with the verified raw-log core.

\subsection{Blockchain graphs and financial forensics}
Public-ledger learning has long used address and transaction graphs. BlockSci provides scalable blockchain analysis infrastructure \citep{Kalodner2017}; Elliptic and later extensions demonstrate graph learning for anti-money-laundering and financial forensics \citep{Weber2019,Bellei2024,ElmougyLiu2023}; and Ethereum-specific network research shows that temporal and multiplex edge information can improve representation and prediction \citep{Lin2020EthereumNetwork}. Dynamic graph models such as GraphSAGE, graph attention, Temporal Graph Networks, and TGAT provide natural comparison families when the learning object is an address--contract--transfer graph \citep{Hamilton2017,Velickovic2018,Rossi2020,Xu2020}. ETH-TraceBench differs in unit of analysis: it keeps the event stream \emph{inside} each transaction as the core object, while graph views remain complementary representations for future comparisons.

\subsection{Self-supervised learning for logs and structured event streams}
Sequence and masked-reconstruction methods provide a second natural comparison family. Transformers and BERT establish generic attention and masked-token learning \citep{Vaswani2017,Devlin2019}; DeepLog and LogBERT adapt sequential and masked-language modeling to system logs \citep{Du2017,Guo2021}. On Ethereum, BERT4ETH pretrains a Transformer over transaction sequences for account representation \citep{Hu2023BERT4ETH}, while more recent systems combine transaction-language modeling with graph structure \citep{TLMG4Eth2024,LMAE4Eth2025}. The empirical evaluation therefore includes a standard supervised Transformer over exact raw-log order in addition to aggregate and bag-of-symbol baselines. The ordered-versus-shuffled ablation tests whether chronological event order adds predictive value beyond event composition. More elaborate masked-event or flow-aware architectures remain future modeling directions and are not part of the current empirical claim.

\subsection{Adjacent DeFi datasets and aggregation levels}
Adjacent DeFi datasets operate at different levels of aggregation. DeXposure constructs protocol-level interdependency networks and benchmark tasks across decentralized financial systems \citep{Wu2025DeXposure}; Uniswap transaction-index datasets aggregate activity over time and networks \citep{Chemaya2023UniswapIndices}. Protocol-level and aggregated DeFi datasets are valuable for ecosystem-level dynamics \citep{Wu2025DeXposure,Chemaya2023UniswapIndices}, whereas ETH-TraceBench fixes transaction-level raw-log event streams and evaluates whether representations remain reliable when the protocols, pools, contracts, and event vocabularies generating those streams change. The benchmark is model-agnostic: parsers, bag-of-event models, sequence models, graph models, and future trace-native encoders can be compared if they use the same partitions and report the same hard-transfer and masking conditions.
}

{\color{black}
\section{Benchmark Definition and Evaluation Settings}
\label{sec:problem}

\begin{table}[!htbp]
\centering
\color{black}
\caption{ETH-TraceBench benchmark card.}
\label{tab:benchmark-card}
\small
\setlength{\tabcolsep}{5pt}
\begin{tabularx}{\textwidth}{p{0.29\textwidth}X}
\toprule
Item & Specification \\
\midrule
Raw universe & Ethereum mainnet, Jan. 2021--Dec. 2025; 1.35B transactions with logs; 5.01B raw log rows \\
Positive-label source population & 311.97M task-labelled transaction instances; 2.33B matched raw logs \\
Primary task & DEX swap/trade classification \\
Stress-test task & Lending-liquidation detection \\
Temporal split & Train 2021--2024; validation 2025H1; test 2025H2 \\
Hard evaluation axes & Protocol transfer; pool/infrastructure holdout; symbolic novelty; emitter/topic masking \\
Canonical supervised sample & 911,267 rows; exact common transaction hashes used across reported baseline families \\
Primary metrics & Macro-F1; AUPRC for imbalanced settings; positive recall; paired bootstrap uncertainty \\
\bottomrule
\end{tabularx}
\end{table}

\subsection{Learning object and task labels}
The verified core input is a transaction-level sequence of Ethereum receipt logs ordered by \texttt{log\_index}. Ethereum's log abstraction records the emitting contract and topic series for each log entry \citep{Wood2014Ethereum}; the benchmark uses the corresponding raw topic and emitting-contract fields. Aggregate trace features such as raw-log count, unique-topic count, and unique-emitter count are derived from the same receipt-log stream. Project, protocol, pool, and curated label-source fields are used to construct labels or evaluation subsets and are not supplied as supervised predictive features in the reported models.

For transaction $i$, the verified core event stream is
\begin{equation}
T_i=(e_{i1},\ldots,e_{in_i}), \qquad e_{ik}=(s_{ik},c_{ik},p_{ik}),
\end{equation}
where $s_{ik}$ is the raw \texttt{topic0} event signature, $c_{ik}$ is the log-emitting smart-contract address, and $p_{ik}$ is the within-transaction \texttt{log\_index}. A supervised instance is $(T_i,y_i,t_i,d_i)$, where $y_i$ is the task label, $t_i$ determines the chronological split, and $d_i$ contains protocol/pool metadata used only to construct evaluation subsets. Aggregate counts and symbolic bags are deterministic views of $T_i$; the ordered Transformer consumes the sequence itself. Optional call trees, token flows, decoded arguments, or block-context channels are outside this verified core unless separately materialized and audited.

DEX positives are transaction hashes present in \texttt{crosschain.dex.trades} for Ethereum. Liquidation positives are transaction hashes present in \texttt{ethereum.lending.liquidations}. Canonical negatives are transactions with raw Ethereum logs that are excluded from the corresponding positive source before deterministic supervised sampling. Thus the benchmark does not use a hand-curated set of trivially unrelated negatives. Section~\ref{sec:results} further evaluates hard negatives that share exact training-positive emitter--topic pairs.

\subsection{Five evaluation settings}
The evaluation distinguishes five conditions. \textbf{Temporal test} means training on 2021--2024, model selection on 2025H1, and final evaluation on 2025H2. \textbf{Protocol transfer} means evaluating DEX positives from project/protocol combinations absent from training. \textbf{Pool holdout} means evaluating positives associated with pool or liquidity-infrastructure identifiers absent from training. \textbf{Symbolic novelty} means evaluating transactions containing an event topic or emitter--topic pair absent from the training vocabulary. \textbf{Masking} means replacing emitter identity, topic identity, or both before feature hashing/model input construction.

The protocol and pool conditions separate two forms of infrastructure change that are often conflated. A new pool can instantiate an already familiar protocol mechanism; a new protocol can introduce a new symbolic and execution regime. The empirical results show that this distinction matters: protocol novelty produces substantially larger degradation than pool novelty.

\subsection{Participant identity and emitter identity}
The reported predictive inputs do not contain transaction initiator, trader, borrower, liquidator, signer, or user-wallet identity. An ``emitter'' is \texttt{ethereum.raw.logs.address}: the smart contract that emitted a log. Consequently, a participant-disjoint user split would not remove an input feature available to the reported models. The relevant identity-shortcut test is emitter-contract masking, which is reported directly in Section~\ref{sec:results}.

\subsection{Why the two tasks are useful}
DEX classification is the main large-scale task because the same broad economic function is implemented through diverse automated-market-maker protocols, routers, pools, versions, and event vocabularies \citep{Xu2023DEXSoK}. It therefore supports naturally occurring transfer tests at substantial scale. Liquidation detection is used differently: it is a rare-event stress test with stronger event-signature concentration and a naturally imbalanced 2025H2 test set. The two tasks are not claimed to exhaust Ethereum behavior; they provide high-coverage, independently sourced anchors for studying whether event-stream representations remain reliable when the underlying infrastructure changes.
}

{\color{black}
\section{Data and Benchmark Construction}
\label{sec:data}

\subsection{Ethereum raw-log universe}
The empirical setting is Ethereum mainnet from January 2021 through December 2025. The raw-log audit covers 1,346,806,238 transactions that emit at least one log and 5,013,719,926 raw log rows. The distribution is highly skewed: transactions with three or more logs account for 80.6\% of all raw-log rows, while a small tail of transactions with at least 51 logs contributes 738.7 million rows. Appendix~\ref{app:extraction-scripts} documents the operational extraction workflow and integrity checks.

\begin{table}[!htbp]
\centering
\color{black}
\caption{Raw Ethereum log universe, 2021--2025.}
\label{tab:raw-universe}
\small
\begin{tabular}{lrrr}
\toprule
Event-complexity bin & Transactions & Raw log rows & Share of raw logs \\
\midrule
1--2 logs & 859,079,266 & 974,119,684 & 19.4\% \\
3--10 logs & 426,997,616 & 2,320,754,180 & 46.3\% \\
11--50 logs & 57,239,096 & 980,142,273 & 19.5\% \\
51+ logs & 3,490,260 & 738,703,789 & 14.7\% \\
\midrule
Total & 1,346,806,238 & 5,013,719,926 & 100.0\% \\
\bottomrule
\end{tabular}
\end{table}

\subsection{Label sources, negatives, and benchmark scale}
The supervised task labels come from two curated on-chain sources. DEX positives are transaction hashes present in \texttt{crosschain.dex.trades}; liquidation positives are transaction hashes present in \texttt{ethereum.lending.liquidations}. The two source tables provide high-precision programmatic task labels rather than a complete taxonomy of Ethereum behavior. The label design follows the weak-supervision view that programmatic sources should be documented and audited rather than treated as exhaustive ground truth by assumption \citep{Ratner2016DataProgramming,Ratner2018Snorkel}. Confident-learning work likewise motivates explicit checks for likely label error and uncertainty \citep{Northcutt2021ConfidentLearning}. Appendix~\ref{app:weak-label-rules} records the exact source-label rules, and Appendix~\ref{app:source-audit} documents the completed source-concordance and event-signature audit. For each task, candidate negatives are Ethereum transactions with raw logs that are not present in the corresponding positive source. The canonical supervised extraction then deterministically samples from these positive and negative pools under the fixed temporal partitions.

Negative construction is important for interpreting the high aggregate scores. Broad source-excluded negatives make the aggregate task partly recoverable from simple execution complexity, but they are not the only negative regime reported. We separately identify 2025H2 negatives that share at least one exact emitter--topic pair with a 2021--2024 training positive and report false-positive rates on this infrastructure-adjacent hard-negative slice.

\begin{table}[!htbp]
\centering
\color{black}
\caption{Positive-label source population.}
\label{tab:benchmark-scale}
\small
\begin{tabular}{lrrr}
\toprule
Task & Labelled transactions & Matched raw logs & Projects \\
\midrule
DEX event classification & 311.87M & 2.33B & 55 \\
Liquidation detection & 105,952 & 2.45M & 21 \\
\bottomrule
\end{tabular}
\end{table}

\subsection{Temporal splits and raw-log coverage}
Training uses 2021--2024, validation uses January--June 2025, and testing uses July--December 2025. The forward split prevents future protocol deployments, future pools, and future event vocabularies from entering supervised training. The 2025H2 DEX population is also compositionally richer than training: average logs per positive transaction rise from 7.05 to 9.32, unique topics from 4.29 to 4.73, and unique emitters from 3.75 to 4.53. Raw-log coverage is nearly complete: 0.035\% of DEX-labelled 2025H2 transactions lack a matched raw log, while all liquidation-labelled test transactions have matched logs. The exact temporal and diagnostic manifests are summarized in Appendix~\ref{app:splits}.

\begin{table}[!htbp]
\centering
\color{black}
\caption{Positive-label population by temporal split.}
\label{tab:benchmark-splits}
\small
\setlength{\tabcolsep}{4pt}
\begin{tabular}{llrrrrr}
\toprule
Task & Split & Transactions & Logs & Logs/tx & Topics/tx & Emitters/tx \\
\midrule
DEX & Train 2021--2024 & 233.41M & 1.65B & 7.05 & 4.29 & 3.75 \\
DEX & Validation 2025H1 & 37.38M & 300.84M & 8.05 & 4.49 & 4.20 \\
DEX & Test 2025H2 & 41.08M & 383.02M & 9.32 & 4.73 & 4.53 \\
Liquidation & Train 2021--2024 & 69,609 & 1.50M & 21.55 & 9.77 & 7.68 \\
Liquidation & Validation 2025H1 & 23,264 & 561,222 & 24.12 & 9.89 & 8.77 \\
Liquidation & Test 2025H2 & 13,079 & 389,513 & 29.78 & 10.55 & 9.41 \\
\bottomrule
\end{tabular}
\end{table}

\subsection{Protocol and pool/infrastructure shift}
Several project/protocol combinations are absent from 2021--2024 training and first appear in 2025. Within the full positive-label population, Uniswap v4 has no training-period transactions, then 1,851,562 validation transactions and 6,600,656 test transactions. Ekubo v1 likewise has no training-period transactions, then 284,549 validation and 1,972,900 test transactions. Together with smaller new project/protocol combinations, protocols absent from training account for 8,618,559 labelled DEX test transactions in the population manifest.

Pool and liquidity-infrastructure turnover is even broader, but the full manifest is an \emph{identifier-incidence} table rather than a distinct-transaction table: a multi-pool or multi-route transaction can contribute to more than one identifier category. Table~\ref{tab:population-holdout} therefore reports transaction--identifier incidences explicitly. We use these full-scale counts to characterize infrastructure turnover and exact transaction-hash subsets for the canonical model evaluations in Section~\ref{sec:results}. Appendix~\ref{app:splits} records the split/holdout manifests, and Appendix~\ref{app:shift-artifacts} records the shortcut and hard-transfer artifacts used to regenerate the restricted evaluations.

\begin{table}[!htbp]
\centering
\color{black}
\caption{Full-scale DEX protocol and pool/liquidity-infrastructure shift. Pool figures are transaction--identifier incidences, not necessarily distinct transactions.}
\label{tab:population-holdout}
\small
\resizebox{\textwidth}{!}{%
\begin{tabular}{llrr}
\toprule
Shift object & Status & Identifiers / projects & 2025H2 mass \\
\midrule
Protocol & Uniswap v4 (absent from train) & 1 project/protocol & 6,600,656 tx \\
Protocol & Ekubo v1 (absent from train) & 1 project/protocol & 1,972,900 tx \\
Protocol & All project/protocol combinations absent from train & -- & 8,618,559 tx \\
\midrule
Pool/infrastructure & Seen in train and test & 55,167 identifiers & 32.8M incidences \\
Pool/infrastructure & First seen in validation, also in test & 8,674 identifiers & 13.4M incidences \\
Pool/infrastructure & Test-only identifier & 55,559 identifiers & 12.8M incidences \\
\bottomrule
\end{tabular}%
}
\end{table}

\subsection{Shortcut diagnostics}
Generic event signatures are strongly concentrated. After excluding common token contracts, the most frequent remaining topic appears in 80.7\% of DEX-labelled test transactions and 95.2\% of liquidation-labelled test transactions; the dominant topic is the ERC-20 Transfer signature. After additionally excluding common Transfer, Approval, Swap, Sync, and related generic topics, the top-topic share falls to 14.35\% for DEX, leaving 6,783 topics and 97,084 emitting contracts. Liquidation remains more concentrated: the top remaining topic appears in 53.53\% of positives across 434 topics and 1,797 emitting contracts. The concentration diagnostics motivate the symbolic-novelty and masking evaluations rather than serving as performance evidence by themselves. Their supporting manifests and completed hard-transfer artifacts are listed in Appendix~\ref{app:shift-artifacts}.

\begin{table}[!htbp]
\centering
\color{black}
\caption{Event-topic shortcut diagnostics in the 2025H2 labelled test populations.}
\label{tab:shortcut-population}
\small
\begin{tabular}{lrrrr}
\toprule
Task & Top-topic share before & Top-topic share after & Topics after & Emitters after \\
\midrule
DEX swap/trade & 80.7\% & 14.35\% & 6,783 & 97,084 \\
Liquidation & 95.2\% & 53.53\% & 434 & 1,797 \\
\bottomrule
\end{tabular}
\end{table}

\subsection{Leakage, label-source separation, and validity controls}
The benchmark separates label construction, model input construction, split assignment, and model selection. Chronological assignment prevents future protocols and pools from entering supervised training through random splits; transaction hashes are unique evaluation keys; decoded source-table fields such as project, protocol, and liquidation metadata are used for labels or evaluation strata rather than predictive inputs; and masking is applied before hashing or model input construction so masked identities cannot be reintroduced through downstream tokenization. Model and threshold selection use 2025H1 only, with 2025H2 reserved for final reporting. Exact hash joins verify same-instance comparisons across the principal baseline families.

Raw-log completeness is audited separately from model error. In 2025H2, only 14,354 of 41,076,269 DEX-labelled transactions have no matched raw log (0.035\%), while all 13,079 liquidation-labelled transactions have matched logs. The participant-identity audit additionally confirms that the reported models do not receive transaction initiator, trader, borrower, liquidator, signer, or user-wallet identity; address-like predictive fields refer to log-emitting contracts. Together, the controls bound what the reported results can and cannot be interpreted as learning. Appendix~\ref{app:reproducibility} summarizes the reproducibility checks, while Appendix~\ref{app:baseline-fairness} records the common split/model-selection controls applied across baseline families.

}

{\color{black}
\section{Tasks, Canonical Samples, and Evaluation Protocol}
\label{sec:evaluation}

\subsection{Canonical supervised sample and same-instance policy}
All reported model comparisons use the same canonical 911,267-row supervised feature universe. The DEX test set contains exactly 85,000 transactions (42,500 positive and 42,500 negative). The liquidation test set contains 143,861 transactions (13,079 positive and 130,782 negative), preserving the natural rare-event imbalance of the extracted 2025H2 liquidation benchmark. Exact transaction-hash checks confirm one-to-one alignment of the canonical DEX test set across TraceStats-GB, TopicEmitterTrace-SGD, TopicEmitterHashMLP, and the ordered/shuffled Transformer predictions.

\begin{table}[!htbp]
\centering
\color{black}
\caption{Canonical supervised evaluation sample.}
\label{tab:canonical-sample}
\small
\begin{tabular}{llrrr}
\toprule
Task & Split & $N$ & Positives & Positive rate \\
\midrule
DEX & Train 2021--2024 & 420,124 & 249,984 & 0.595 \\
DEX & Validation 2025H1 & 85,000 & 42,500 & 0.500 \\
DEX & Test 2025H2 & 85,000 & 42,500 & 0.500 \\
Liquidation & Train 2021--2024 & 130,754 & 69,609 & 0.532 \\
Liquidation & Validation 2025H1 & 46,528 & 23,264 & 0.500 \\
Liquidation & Test 2025H2 & 143,861 & 13,079 & 0.091 \\
\bottomrule
\end{tabular}
\end{table}

The 50/50 DEX validation and test sets are controlled diagnostic samples, not estimates of deployment prevalence. Section~\ref{sec:results} therefore reports a separate population-prevalence and representativeness audit and reweights the fixed predictions to the measured 2025H2 DEX prior as a sensitivity analysis.

\subsection{Reported model families}
\textbf{TraceStats-GB} uses only raw-log count, unique-topic count, and unique-emitter count. \textbf{TopicEmitterTrace-SGD} uses hashed topic, emitter, and emitter--topic symbolic bags plus the three trace counts. \textbf{TopicEmitterHashMLP} embeds hashed topic, emitter, and emitter--topic tokens, mean-pools the symbolic set, and applies a small MLP. \textbf{OrderedEventTransformer} is a standard supervised Transformer over exact \texttt{log\_index}-ordered emitter--topic events. Its deterministic shuffled-order ablation contains the same retained events for each transaction and differs only in order.

The symbolic topic/emitter inputs are bags, not chronological sequences. Only the ordered-event extraction used by the Transformer preserves event order. Separating unordered symbolic bags from ordered sequences makes the Transformer comparison a direct diagnostic of whether order adds information beyond symbolic composition. Appendix~\ref{app:complete-baselines} preserves the complete canonical trace-count and symbolic baseline families (LR/RF/GB and TopicBag/EmitterBag/TopicEmitterHash/TopicEmitterTrace) for validation and test, while Appendix~\ref{app:baseline-fairness} records the common evaluation controls. The main text can therefore focus on representative models without omitting the broader baseline evidence.

\subsection{Model selection, metrics, and uncertainty}
Model selection uses only training data and 2025H1 validation. For the masking and Transformer experiments, the decision threshold is selected on 2025H1 to maximize macro-F1 and then frozen for 2025H2. No threshold is retuned inside the test bootstrap. Macro-F1 is the primary balanced summary, while AUPRC is used for rare-event and ranking quality because precision--recall analysis is more informative than AUROC when positives are sparse \citep{Davis2006PRROC,Saito2015PR}; positive recall is reported where useful. Confidence intervals for ranking metrics require care \citep{Boyd2013AUPRC}; our principal uncertainty comparison is therefore a paired transaction bootstrap on exact common test instances. Neural experiments use three seeds. Transaction-level paired bootstrap intervals are calculated on the exact common test instances; training-seed standard deviations are reported separately so that transaction uncertainty is not misinterpreted as training-run uncertainty. Appendix~\ref{sec:app-artifact-release-tables} specifies the prediction-file schema and reporting fields required to regenerate these comparisons.

\subsection{Natural prevalence and prior-shift sensitivity}
The 2025H2 raw-log universe contains 182,557,827 distinct Ethereum transactions with logs and 41,081,186 DEX-labelled transactions, corresponding to a raw prevalence of approximately 22.50\%. The benchmark-joined DEX count gives a nearly identical 22.5004\% prior, which we use for the sensitivity calculation. The analysis reweights the fixed canonical predictions to a population label prior; it does not imply that every one of the 182.6 million transactions was scored by every model.
}

{\color{black}
\section{Results: Aggregate Performance, Hard Transfer, and Validity Audits}
\label{sec:results}

\subsection{Aggregate canonical baselines}
Table~\ref{tab:canonical-baselines} reports the canonical 2025H2 test results. Ordinary temporal classification is highly predictable from simple trace and symbolic features. On DEX, TraceStats-GB reaches 0.953 macro-F1 and TopicEmitterHashMLP reaches 0.959. Liquidation is also highly predictable from symbolic features despite its 9.1\% positive rate. The aggregate values serve as diagnostics of available signal; benchmark difficulty is assessed in the hard-transfer and controlled-input conditions that follow.

\begin{table}[!htbp]
\centering
\color{black}
\caption{Canonical 2025H2 baseline results on exact common evaluation instances. Values are means across three seeds.}
\label{tab:canonical-baselines}
\small
\setlength{\tabcolsep}{4pt}
\begin{tabular}{llrrrr}
\toprule
Task & Model & $N$ & Macro-F1 & AUPRC & Positive recall \\
\midrule
DEX & TraceStats-GB & 85,000 & 0.9528 & 0.9610 & 0.9733 \\
DEX & TopicEmitterTrace-SGD & 85,000 & 0.9239 & 0.9670 & 0.8909 \\
DEX & TopicEmitterHashMLP & 85,000 & \textbf{0.9595} & \textbf{0.9939} & 0.9424 \\
DEX & Ordered Transformer & 85,000 & 0.9427 & 0.9882 & 0.9000 \\
DEX & Shuffled-order Transformer & 85,000 & 0.9529 & 0.9927 & 0.9226 \\
\midrule
Liquidation & TraceStats-GB & 143,861 & 0.8470 & 0.8205 & 0.9703 \\
Liquidation & TopicEmitterTrace-SGD & 143,861 & 0.9430 & 0.9389 & 0.9662 \\
Liquidation & TopicEmitterHashMLP & 143,861 & 0.9382 & 0.9645 & 0.9818 \\
Liquidation & Ordered Transformer & 143,861 & 0.9472 & 0.9775 & 0.9867 \\
Liquidation & Shuffled-order Transformer & 143,861 & 0.9479 & 0.9779 & 0.9865 \\
\bottomrule
\end{tabular}
\end{table}

For DEX macro-F1, paired transaction bootstrap 95\% intervals are [0.9514, 0.9542] for TraceStats-GB, [0.9222, 0.9258] for TopicEmitterTrace-SGD, [0.9581, 0.9608] for TopicEmitterHashMLP, [0.9412, 0.9441] for the ordered Transformer, and [0.9516, 0.9541] for the shuffled-order Transformer. The intervals quantify test-instance uncertainty; seed-to-seed variation is reported separately for the neural models.

\subsection{Symbolic novelty exposes shortcut-sensitive transfer gaps}
Symbolic novelty isolates a failure mode distinct from protocol and pool shift. For each task, training-set vocabularies are built from event topics and exact emitter--topic pairs. Test transactions are then separated according to whether all symbolic elements were seen in 2021--2024 training or at least one element is novel.

\begin{table}[!htbp]
\centering
\color{black}
\caption{Symbolic-novelty audit on the 2025H2 test split. ``Unseen'' means at least one topic or emitter--topic pair is absent from the task-specific training vocabulary.}
\label{tab:verified-ood-slice-audit}
\scriptsize
\setlength{\tabcolsep}{3pt}
\resizebox{\textwidth}{!}{%
\begin{tabular}{lllrrrrrrr}
\toprule
Task & Model & Slice & Seen $N$ & Unseen $N$ & Seen MF1 & Unseen MF1 & Gap & Seen AUPRC & Unseen AUPRC \\
\midrule
DEX & TETrace-SGD & Topic & 69,228 & 15,772 & 0.950 & 0.772 & $-0.178$ & 0.975 & 0.948 \\
DEX & TETrace-SGD & Emitter--topic pair & 44,364 & 40,636 & 0.960 & 0.860 & $-0.100$ & 0.965 & 0.970 \\
DEX & TopicBag-SGD & Topic & 69,228 & 15,772 & 0.977 & 0.655 & $-0.322$ & 0.996 & 0.979 \\
DEX & TopicBag-SGD & Emitter--topic pair & 44,364 & 40,636 & 0.983 & 0.834 & $-0.149$ & 0.994 & 0.988 \\
Liquidation & TETrace-SGD & Topic & 120,455 & 23,406 & 0.973 & 0.878 & $-0.095$ & 0.985 & 0.878 \\
Liquidation & TETrace-SGD & Emitter--topic pair & 85,879 & 57,982 & 0.989 & 0.908 & $-0.081$ & 0.997 & 0.899 \\
Liquidation & TopicBag-SGD & Topic & 120,455 & 23,406 & 0.953 & 0.912 & $-0.041$ & 0.968 & 0.975 \\
Liquidation & TopicBag-SGD & Emitter--topic pair & 85,879 & 57,982 & 0.971 & 0.917 & $-0.054$ & 0.978 & 0.955 \\
\bottomrule
\end{tabular}%
}
\end{table}

The DEX gap is especially large: TopicBag-SGD falls from 0.977 to 0.655 macro-F1 on unseen-topic transactions, while TopicEmitterTrace-SGD falls from 0.950 to 0.772. On unseen emitter--topic pairs the corresponding changes are 0.983 to 0.834 and 0.960 to 0.860. Liquidation is less brittle but still shows an emitter--topic novelty gap for TopicEmitterTrace-SGD (0.989 to 0.908). The gaps isolate a different failure mode from protocol novelty: public symbols can fail to transfer even when the label family is unchanged.

\subsection{Protocol novelty is substantially harder than pool novelty}
The main benchmark result is the gap between aggregate future-period performance and strict protocol transfer. Table~\ref{tab:hard-transfer} uses the same 42,500 canonical DEX negatives for every challenge subset and restricts the positive side to the stated transfer condition. Strict unseen-protocol performance drops sharply for all three baseline families. By contrast, strict unseen-pool performance remains much closer to the full test score.

\begin{table}[!htbp]
\centering
\color{black}
\caption{DEX hard-transfer macro-F1 on the canonical 2025H2 test set. Challenge subsets retain the same 42,500 canonical negatives and restrict the positive set.}
\label{tab:hard-transfer}
\small
\setlength{\tabcolsep}{4pt}
\begin{tabular}{lrrrrr}
\toprule
Model & Full test & Strict unseen protocol & Uniswap v4 & Ekubo v1 & Strict unseen pool \\
\midrule
TraceStats-GB & 0.9528 & 0.7944 & 0.8513 & 0.7689 & 0.9268 \\
TopicEmitterTrace-SGD & 0.9239 & 0.7434 & 0.8339 & 0.7906 & 0.8973 \\
TopicEmitterHashMLP & 0.9595 & 0.7658 & 0.8657 & 0.8432 & 0.9346 \\
\bottomrule
\end{tabular}
\end{table}

The strict unseen-protocol subset contains 4,449 positives. The named protocol subsets contain 6,701 Uniswap v4 positives and 2,051 Ekubo v1 positives. Paired bootstrap analysis estimates full-to-strict-protocol macro-F1 degradation of approximately $-0.158$ for TraceStats-GB, $-0.180$ for TopicEmitterTrace-SGD, and $-0.194$ for TopicEmitterHashMLP. The corresponding strict unseen-pool degradation is only about $-0.025$ to $-0.027$. Thus a high aggregate temporal score does not imply robustness to a new protocol regime, and protocol novelty is empirically different from a new pool within more familiar infrastructure.

\subsection{Masking: contract-emitter identity alone does not explain DEX performance}
Table~\ref{tab:masking-results} retrains TopicEmitterTrace-SGD under four input views and uses thresholds selected on 2025H1. ``Emitter masked'' replaces the log-emitting contract identity; ``topic masked'' replaces event-topic identity; the joint condition removes both symbolic identity channels before hashing.

\begin{table}[!htbp]
\centering
\color{black}
\caption{Masking robustness for TopicEmitterTrace-SGD. Test thresholds are selected on 2025H1 and frozen for 2025H2.}
\label{tab:masking-results}
\small
\begin{tabular}{llrrr}
\toprule
Task & Input view & Macro-F1 & AUPRC & Positive recall \\
\midrule
DEX & Full & 0.9327 & 0.9670 & 0.9157 \\
DEX & Emitter masked & \textbf{0.9442} & 0.9908 & 0.9019 \\
DEX & Topic masked & 0.9331 & 0.9616 & 0.9168 \\
DEX & Emitter + topic masked & 0.9165 & 0.8572 & 0.9970 \\
\midrule
Liquidation & Full & 0.9439 & 0.9389 & 0.9629 \\
Liquidation & Emitter masked & \textbf{0.9604} & 0.9613 & 0.9794 \\
Liquidation & Topic masked & 0.9402 & 0.9197 & 0.9597 \\
Liquidation & Emitter + topic masked & 0.7741 & 0.6377 & 0.9905 \\
\bottomrule
\end{tabular}
\end{table}

\noindent\textit{Note.} Table~\ref{tab:masking-results} is a self-contained retraining ablation. The ``Full'' row is the unmasked control regenerated under the same masking-experiment training and validation-threshold protocol as the three masked views. It therefore differs slightly from the canonical TopicEmitterTrace-SGD row in Table~\ref{tab:canonical-baselines}, which comes from the canonical baseline run. Masking effects are interpreted relative to the within-table Full control rather than as a cross-table difference.

Emitter masking alone does not reduce DEX performance and therefore contradicts a simple contract-address memorization explanation. Topic masking alone also preserves most DEX performance. The substantial deterioration occurs when both symbolic identity channels are removed, especially for liquidation. The masking pattern is consistent with multiple redundant public signals rather than dependence on one repeated actor or one contract address.

\subsection{A standard sequence model: true event order provides no consistent advantage}
The ordered-event extraction reconstructs each transaction's raw-log events in exact \texttt{log\_index} order. A standard supervised Transformer receives emitter/topic pairs with a maximum retained length of 128 events. The shuffled condition applies a deterministic transaction-specific permutation to the same retained events, so the two conditions differ only in order.

\begin{table}[!htbp]
\centering
\color{black}
\caption{Ordered versus shuffled event sequence Transformer, mean $\pm$ seed SD over three seeds.}
\label{tab:transformer-order}
\small
\begin{tabular}{llrr}
\toprule
Task & Condition & Macro-F1 & AUPRC \\
\midrule
DEX & Ordered & $0.9427 \pm 0.0047$ & $0.9882 \pm 0.0057$ \\
DEX & Shuffled & $0.9529 \pm 0.0123$ & $0.9927 \pm 0.0022$ \\
Liquidation & Ordered & $0.9472 \pm 0.0137$ & $0.9775 \pm 0.0122$ \\
Liquidation & Shuffled & $0.9479 \pm 0.0267$ & $0.9779 \pm 0.0103$ \\
\bottomrule
\end{tabular}
\end{table}

For DEX, the paired ordered-minus-shuffled macro-F1 difference is $-0.0102$ with a transaction-bootstrap 95\% interval of [$-0.0108$, $-0.0096$]. For liquidation the difference is $-0.0007$ with interval [$-0.0016$, 0.0002]. We do not interpret the DEX result as evidence that shuffling is intrinsically beneficial; rather, across the two tasks there is no consistent evidence that true within-transaction event order adds predictive value beyond the retained event multiset. The ablation therefore qualifies claims that high performance reflects sophisticated chronological event modeling.

\subsection{Hard negatives expose infrastructure-adjacent errors}
For each task, we build emitter, topic, and exact emitter--topic support sets from \emph{training positives only}. A 2025H2 labelled negative is a pair-overlap hard negative if it contains at least one exact emitter--topic pair previously observed in a 2021--2024 positive. Pair overlap is common rather than exceptional: 32,470 of 42,500 DEX negatives (76.4\%) and 99,476 of 130,782 liquidation negatives (76.1\%) satisfy it.

\begin{table}[!htbp]
\centering
\color{black}
\caption{False-positive rates on symbolic-overlap hard negatives versus negatives with no exact training-positive pair overlap.}
\label{tab:hard-negatives}
\small
\begin{tabular}{llrr}
\toprule
Task & Model & Pair-overlap FPR & No-pair-overlap FPR \\
\midrule
DEX & TraceStats-GB & 0.0803 & 0.0265 \\
DEX & TopicEmitterTrace-SGD & 0.0548 & 0.0041 \\
DEX & TopicEmitterHashMLP & 0.0245 & 0.0201 \\
DEX & Ordered Transformer & 0.0171 & 0.0055 \\
\midrule
Liquidation & TraceStats-GB & 0.0874 & 0.0084 \\
Liquidation & TopicEmitterTrace-SGD & 0.0245 & 0.0006 \\
Liquidation & TopicEmitterHashMLP & 0.0275 & 0.0077 \\
Liquidation & Ordered Transformer & 0.0228 & 0.0089 \\
\bottomrule
\end{tabular}
\end{table}

The higher FPRs show that negatives sharing training-positive symbolic infrastructure are more confusable. The stress test also clarifies why the strong aggregate task remains useful: it separates easy broad negatives from infrastructure-adjacent cases without changing the underlying source-table label definition.

\subsection{Natural prevalence and sample representativeness}
The balanced DEX test is intended for controlled model comparison, not prevalence estimation. Representativeness is therefore assessed against the full 2025H2 population. Among 182,557,827 distinct Ethereum transactions with logs, 41,081,186 are DEX-labelled, a raw prevalence of approximately 22.50\%. The exact canonical positive sample tracks the DEX population closely: total-variation distance is 0.0042 by month, 0.0031 by project, and 0.0065 by protocol; the observed project and protocol categories cover more than 99.98\% of the corresponding population mass. The canonical negative month distribution has total-variation distance 0.0021 from the full non-DEX logged-transaction population. Mean differences in logs, unique topics, and unique emitters per positive transaction are all below 0.4\% relative to the full DEX test reference.

\begin{table}[!htbp]
\centering
\color{black}
\caption{DEX prior-shift sensitivity using the measured 2025H2 benchmark-matched prevalence of 22.5004\%. Values are obtained by reweighting the fixed canonical score/error distributions; this is not a full-population scoring run.}
\label{tab:natural-prior}
\small
\begin{tabular}{lrrrr}
\toprule
Model & Balanced MF1 & Prior-reweighted MF1 & Balanced AUPRC & Prior-reweighted AUPRC \\
\midrule
TraceStats-GB & 0.9528 & 0.9218 & 0.9610 & 0.8794 \\
TopicEmitterTrace-SGD & 0.9239 & 0.9183 & 0.9670 & 0.9013 \\
TopicEmitterHashMLP & 0.9595 & 0.9557 & 0.9939 & 0.9810 \\
Ordered Transformer & 0.9427 & 0.9509 & 0.9882 & 0.9699 \\
Shuffled-order Transformer & 0.9529 & 0.9562 & 0.9927 & 0.9789 \\
\bottomrule
\end{tabular}
\end{table}

\subsection{Source-concordance and event-signature validity audit}
A deterministic 400-transaction audit sample is stratified across July--December 2025. For each task it contains 100 labelled positives, 50 ordinary negatives with no exact training-positive pair overlap, and 50 hard negatives with pair overlap. Every transaction is independently re-queried against the task source and \texttt{ethereum.raw.logs}. Appendix~\ref{app:source-audit} provides the audit design and distinguishes the automated source/signature checks from the optional future human-review template.

\begin{table}[!htbp]
\centering
\color{black}
\caption{Automated label/source and event-signature audit.}
\label{tab:label-audit}
\small
\begin{tabular}{lrr}
\toprule
Audit check & Result & Rate \\
\midrule
Source-table membership concordance & 400 / 400 & 100\% \\
Raw-log presence & 400 / 400 & 100\% \\
Canonical vs re-queried raw-log count agreement & 400 / 400 & 100\% \\
Positive transactions containing an audited task-positive signature & 200 / 200 & 100\% \\
Negative transactions containing an audited task-positive signature & 0 / 200 & 0\% \\
\bottomrule
\end{tabular}
\end{table}

For DEX, the positive-source rows yield 11 observed trade-event topic signatures; for liquidation, the source table yields six liquidation-event topics. All sampled positives contain at least one corresponding signature and none of the sampled negatives does, including all 100 symbolic-overlap hard negatives. No independent double-coded human semantic annotation was collected for this version; the audit therefore supports source/signature concordance and should not be interpreted as inter-annotator validation. The signature evidence strengthens the source-concordance result without claiming exhaustive human semantic annotation or complete coverage of every possible protocol-specific signature.

\subsection{Interpretation}
High aggregate scores coexist with substantial degradation under protocol novelty and shortcut controls. Ethereum exposes stable, public, highly predictive symbols, but performance falls when those symbols and mechanisms change: protocol novelty produces large losses, symbolic masking reveals redundant shortcut channels, infrastructure-adjacent negatives concentrate false positives, and chronological order itself provides little incremental signal in a standard Transformer. ETH-TraceBench therefore evaluates whether a representation survives naturally occurring changes in the data-generating infrastructure, with the full future-period score serving as only one part of that assessment.
}

{\color{black}
\section{Release and Reproducibility}
\label{sec:release}
The release is organized around artifacts that make the reported comparisons regenerable. Construction artifacts record the raw-log universe, task labels, temporal splits, protocol/pool manifests, topic diagnostics, and raw-log coverage. Model artifacts record configuration, seed, exact transaction hash, task, split, condition, true label, score, and predicted label. The release additionally archives canonical holdout manifests, masking outputs, ordered-sequence counts, Transformer predictions, paired-bootstrap outputs, prevalence/representativeness diagnostics, hard-negative manifests, and label-audit evidence. Appendix~\ref{sec:app-artifact-release-tables} details the artifact schemas and package components, Appendix~\ref{app:reproducibility} gives the reproducibility checklist, and Appendix~\ref{app:repo-structure} gives the repository layout.

All reported model comparisons use fixed temporal partitions and 2025H1-only model/threshold selection. The 2025H2 test set is not used for tuning. Same-instance prediction files allow paired comparisons and transaction-level bootstrap intervals to be regenerated. Full raw blockchain mirrors need not be redistributed because the extraction queries and manifests define reconstruction from public chain-derived data; smaller supervised artifacts and exact hashes/keys fix the reported evaluation objects.

For public release, transaction hashes may be replaced by stable salted keys when raw identifiers are unnecessary for metric regeneration. Any restricted derived artifacts should use encryption at rest and in transit together with access controls. Encryption is a storage/access safeguard rather than an anonymization claim: public-chain activity remains inherently observable. The release should not include wallet-to-person mappings and should sanitize or delay examples that could facilitate exploitation of an active vulnerability.
}

{\color{black}
\section{Blockchain Monitoring Implications}
The evidence supports a monitoring interpretation centered on robustness under infrastructure change. Aggregate trace counts and public symbolic features are powerful predictors, so a model that performs well on an ordinary future-period test may still be relying on infrastructure-specific regularities. The strict protocol-transfer results show where this matters most: new protocol regimes are substantially harder than new pools within more familiar infrastructure. For practical monitoring, the benchmark therefore encourages reporting the full temporal score together with protocol-transfer, pool-holdout, symbolic-novelty, and masking results rather than treating any one score as sufficient.

The sequence ablation also informs model design. A standard Transformer does not gain a consistent advantage from true raw-log order over a fixed shuffle of the same events. The result does not imply that transaction structure is irrelevant; it shows only that chronological receipt-log order is not the missing ingredient. Richer models may need typed token-flow direction, call structure, state changes, or other channels beyond the verified core raw-log view, and such claims should be tested on the same hard-transfer conditions.
}

{\color{black}
\section{Limitations and Responsible Release}
The verified core benchmark uses Ethereum receipt logs. Receipt-log event streams, internal call trees, reconstructed token flows, and full EVM execution paths are distinct data objects, so receipt logs are not described as complete execution traces. The ordered Transformer models \texttt{log\_index} order, not nested call order. Future work can add internal calls or flow channels, but those additions require separate leakage controls and artifact provenance.

The task labels are programmatic labels from curated DEX-trade and lending-liquidation tables. The 400-transaction audit provides complete source-table concordance and event-signature consistency on the deterministic audit sample. No independent double-coded human semantic labels were collected for this version, so the audit is not inter-annotator validation and cannot prove exhaustive coverage of every router, private route, upgrade, or protocol-specific mechanism. The source tables can also change over time, so benchmark versions should retain query dates, hashes, and coverage statements.

The canonical DEX evaluation is deliberately class-balanced for controlled comparison. Natural 2025H2 DEX prevalence among logged transactions is approximately 22.5\%; we therefore report sample-representativeness diagnostics and prior-shift sensitivity separately. Prior reweighting preserves the measured class-conditional score/error distributions and should not be interpreted as a full 182.6M-transaction inference run.

The participant-identity concern is limited by the reported inputs: transaction initiator and user/trader wallet identity are not predictive features in the verified models. Emitter identity refers to the smart contract that produced a log. Nonetheless, prior measurement and privacy work shows that pseudonymous public-ledger activity can sometimes be clustered or linked to real-world entities when combined with external information \citep{Meiklejohn2013,Buterin2024Privacy}. The public package should therefore avoid wallet-to-person mappings, use stable salted keys where exact transaction hashes are unnecessary, encrypt restricted derived artifacts at rest and in transit, enforce access controls on any non-public files, and sanitize or delay examples involving active vulnerabilities. Such controls reduce avoidable disclosure and misuse risk without implying that public-chain activity itself can be made private. Appendix~\ref{app:responsible-release} provides the responsible-release checklist.

The empirical setting is Ethereum mainnet. The evaluation framework should transfer most directly to EVM-compatible chains that expose similar logs and smart-contract execution semantics; non-EVM chains require adapted event representations and split definitions.
}

{\color{black}
\section{Conclusions}
ETH-TraceBench provides a ledger-scale benchmark for Ethereum DeFi raw-log event streams under chronological and infrastructure change. Benchmark difficulty is determined by the hard-transfer conditions, not by the aggregate temporal score alone. Simple trace-count and symbolic models perform very strongly on the canonical 2025H2 temporal test, but strict unseen-protocol evaluation produces large degradation, while strict unseen-pool evaluation is substantially easier. Named new protocols such as Uniswap v4 and Ekubo v1 are also materially harder than the full test.

Shortcut-controlled experiments refine this picture. Emitter-contract masking alone does not reduce DEX performance, whereas jointly removing emitter and topic identity causes a meaningful loss and severely degrades liquidation performance. A standard ordered Transformer provides no consistent advantage over a deterministic shuffle of the same retained events, indicating that chronological raw-log order is not sufficient to explain robust transfer. Hard negatives sharing training-positive symbolic infrastructure concentrate false positives, while the 400-transaction source-concordance and event-signature audit supports the validity of the task labels on ordinary and hard-negative cases.

The benchmark's main contribution is therefore an evaluation object and protocol for distinguishing high in-distribution predictability from transferable event-stream representation. Future models---including richer sequence, flow, graph, or masked-event architectures---should be compared on the same exact instances and should report protocol transfer, pool holdout, symbolic novelty, masking, hard-negative behavior, and uncertainty alongside aggregate temporal performance.
}

\section*{Ethics statement}
{\color{black}
The study uses public blockchain data and does not involve human participants, private off-chain user data, or attempts to deanonymize wallet owners. The release avoids wallet-to-person mappings and treats exact public-chain identifiers as reproducibility fields rather than human identity claims. Restricted derived artifacts, if any, should use encryption and access controls; public examples involving active vulnerabilities should be sanitized or delayed.
}

\section*{Funding}
The research did not receive any specific grant from funding agencies in the public, commercial, or not-for-profit sectors.

\section*{Data availability}
{\color{black}
The study uses public Ethereum blockchain data. The verified package contains extraction/query manifests, temporal and holdout definitions, canonical supervised instance manifests, model configurations, per-seed prediction artifacts for the reported baselines and sequence experiments, masking outputs, paired-bootstrap code/results, prevalence and representativeness diagnostics, hard-negative manifests, and label/source-audit evidence; Appendix~\ref{app:supplement} documents these supplementary components and their status. Full raw blockchain mirrors may be reconstructed from public chain data rather than redistributed. Exact public transaction hashes may be replaced by stable salted keys in public model-artifact packages when direct identifiers are unnecessary for metric regeneration.
}

\section*{Declaration of competing interest}
Declarations of interest: none.

\section*{Acknowledgements}
Omitted for double-blind review.

\appendix

\section{Supplementary Material and Reproducibility Appendix}

{\color{black}
The appendix contains detailed release schemas, complete canonical baseline families, operational and reproducibility material, and the MEM-BL architecture/configuration as a \emph{future reference direction only}. The empirical conclusions rely on the evaluated protocol/pool holdouts, masking runs, ordered/shuffled Transformer, uncertainty analysis, representativeness audit, hard-negative stress test, and source/event-signature validity audits; MEM-BL is retained only as a specification for future modeling work.
}

\label{app:supplement}

Appendix A reports the supplementary materials associated with the extraction, labeling, training, evaluation, auditing, and release workflow. It documents the model configuration, hyperparameter search space, baseline controls, weak-label rules, audit protocol, split manifests, shortcut tests, objective-level ablations, representation diagnostics, runtime scaling, optional external validation, error-analysis templates, responsible-release safeguards, and reproducibility package structure.

\subsection{Complete canonical baseline-family results}
\label{app:complete-baselines}
{\color{black}
Tables~\ref{tab:app-trace-baselines} and \ref{tab:app-symbolic-baselines} report the complete verified trace-count and linear symbolic baseline families on the canonical validation and test instances, regenerated from the archived per-seed prediction files.

\begin{table}[!htbp]
\centering
\color{black}
\caption{Complete canonical trace-count baseline family (means over three seeds).}
\label{tab:app-trace-baselines}
\scriptsize
\begin{tabular}{lllrrrr}
\toprule
Task & Split & Model & Macro-F1 & Pos.-F1 & AUROC & AUPRC \\
\midrule
DEX & Test 2025H2 & TraceStats-GB & 0.953 & 0.954 & 0.975 & 0.961 \\
DEX & Test 2025H2 & TraceStats-LR & 0.864 & 0.861 & 0.943 & 0.910 \\
DEX & Test 2025H2 & TraceStats-RF & 0.949 & 0.951 & 0.971 & 0.954 \\
DEX & Val. 2025H1 & TraceStats-GB & 0.964 & 0.965 & 0.981 & 0.968 \\
DEX & Val. 2025H1 & TraceStats-LR & 0.861 & 0.858 & 0.943 & 0.911 \\
DEX & Val. 2025H1 & TraceStats-RF & 0.960 & 0.961 & 0.978 & 0.963 \\
Liquidation & Test 2025H2 & TraceStats-GB & 0.847 & 0.731 & 0.985 & 0.820 \\
Liquidation & Test 2025H2 & TraceStats-LR & 0.828 & 0.699 & 0.984 & 0.796 \\
Liquidation & Test 2025H2 & TraceStats-RF & 0.843 & 0.724 & 0.986 & 0.831 \\
Liquidation & Val. 2025H1 & TraceStats-GB & 0.931 & 0.930 & 0.988 & 0.984 \\
Liquidation & Val. 2025H1 & TraceStats-LR & 0.921 & 0.921 & 0.983 & 0.975 \\
Liquidation & Val. 2025H1 & TraceStats-RF & 0.929 & 0.928 & 0.988 & 0.984 \\
\bottomrule
\end{tabular}
\end{table}

\begin{table}[!htbp]
\centering
\color{black}
\caption{Complete canonical Ethereum-native symbolic baseline family (means over three seeds).}
\label{tab:app-symbolic-baselines}
\scriptsize
\begin{tabular}{lllrrrr}
\toprule
Task & Split & Model & Macro-F1 & Pos.-F1 & AUROC & AUPRC \\
\midrule
DEX & Test 2025H2 & EmitterBag-SGD & 0.900 & 0.894 & 0.946 & 0.924 \\
DEX & Test 2025H2 & TopicBag-SGD & 0.918 & 0.912 & 0.966 & 0.978 \\
DEX & Test 2025H2 & TEHash-SGD & 0.899 & 0.891 & 0.954 & 0.939 \\
DEX & Test 2025H2 & TETrace-SGD & 0.924 & 0.921 & 0.975 & 0.967 \\
DEX & Val. 2025H1 & EmitterBag-SGD & 0.928 & 0.926 & 0.958 & 0.925 \\
DEX & Val. 2025H1 & TopicBag-SGD & 0.973 & 0.972 & 0.992 & 0.994 \\
DEX & Val. 2025H1 & TEHash-SGD & 0.933 & 0.931 & 0.970 & 0.954 \\
DEX & Val. 2025H1 & TETrace-SGD & 0.949 & 0.949 & 0.982 & 0.973 \\
Liquidation & Test 2025H2 & EmitterBag-SGD & 0.937 & 0.886 & 0.983 & 0.899 \\
Liquidation & Test 2025H2 & TopicBag-SGD & 0.942 & 0.895 & 0.996 & 0.965 \\
Liquidation & Test 2025H2 & TEHash-SGD & 0.943 & 0.895 & 0.986 & 0.945 \\
Liquidation & Test 2025H2 & TETrace-SGD & 0.943 & 0.897 & 0.996 & 0.939 \\
Liquidation & Val. 2025H1 & EmitterBag-SGD & 0.981 & 0.981 & 0.995 & 0.993 \\
Liquidation & Val. 2025H1 & TopicBag-SGD & 0.984 & 0.984 & 0.999 & 0.998 \\
Liquidation & Val. 2025H1 & TEHash-SGD & 0.983 & 0.983 & 0.998 & 0.998 \\
Liquidation & Val. 2025H1 & TETrace-SGD & 0.992 & 0.993 & 0.998 & 0.997 \\
\bottomrule
\end{tabular}
\end{table}
}

\subsection{Future reference architecture: MEM-BL}
\label{app:membl-reference}
{\color{black}
Masked Event Modeling for Blockchain Logs (MEM-BL) is specified here as a \emph{future reference architecture}. The empirical study uses the evaluated baseline, transfer, masking, and sequence results reported above; the MEM-BL specification preserves a possible trace-native modeling direction for future work.

\subsubsection{Event representation and encoder}
The reference design represents a focal transaction as a structured event sequence and optionally augments the verified raw-log core with additional channels when those channels have been materialized with leakage-safe provenance. For event $e_i$ in transaction $T_j$, the proposed embedding is
\begin{equation}
x_i = E_s(s_i)+E_c(c_i)+E_a(a_i)+E_r(r_i)+E_p(p_i)+E_d(d_i)+E_{\Delta}(\Delta_i)+E_x(b(x_i))+E_g(g_i)+W_q q_i,
\end{equation}
where $s_i$ is the event topic or decoded signature, $c_i$ is the emitting contract, $a_i$ is an optional asset identifier, $r_i$ is an optional weak economic-role tag, $p_i$ records position, $d_i$ is optional call depth, $b(x_i)$ is an amount bin, and $q_i$ contains optional continuous covariates such as gas ratios or contract-age features. In the verified benchmark used in the main experiments, only raw-log fields and explicitly derived views are assumed available; the additional channels in this reference equation are future extensions rather than hidden inputs to the reported models.

A transaction-local Transformer maps the event embeddings into contextual event states,
\begin{equation}
h_{ij}=f^{\mathrm{event}}_{\theta}(x_{1:n_j}),
\end{equation}
and attention pooling produces a transaction representation
\begin{equation}
z_j=\sum_{i=1}^{n_j}\alpha_{ij}h_{ij}, \qquad
\alpha_{ij}=\frac{\exp(u^\top \tanh(W h_{ij}))}{\sum_{k=1}^{n_j}\exp(u^\top \tanh(W h_{kj}))}.
\end{equation}
The original design further allowed a leakage-safe block-context encoder and time-stamped address/contract memory, queried only with information observable at or before the focal block.

\subsubsection{Proposed self-supervised objectives}
The original reference protocol corrupted transaction windows and reconstructed missing execution structure with
\begin{equation}
\mathcal{L}=\mathcal{L}_{\mathrm{topic}}+\lambda_a\mathcal{L}_{\mathrm{attr}}+\lambda_c\mathcal{L}_{\mathrm{ctx}}+\lambda_o\mathcal{L}_{\mathrm{ord}}+\lambda_f\mathcal{L}_{\mathrm{flow}}.
\end{equation}
Masked topic prediction reconstructs hidden event signatures or topic hashes,
\begin{equation}
\mathcal{L}_{\mathrm{topic}}=-\sum_{i\in M_s}\log p_{\theta}(s_i\mid \widetilde{W}_j),
\end{equation}
while masked attribute prediction reconstructs available contract, asset, role, or amount-bin attributes,
\begin{equation}
\mathcal{L}_{\mathrm{attr}}=-\sum_{i\in M_a}\sum_{q\in\{c,a,r,b(x)\}}\log p_{\theta}(q_i\mid \widetilde{W}_j).
\end{equation}
A context objective can remove an event span $S_i$ and identify the true span from hard negatives drawn from related protocols or bytecode families,
\begin{equation}
\mathcal{L}_{\mathrm{ctx}}=-\sum_{i\in M_c}
\log \frac{\exp(\mathrm{sim}(z_j,z_{S_i})/\tau)}
{\sum_{S'\in C_i}\exp(\mathrm{sim}(z_j,z_{S'})/\tau)}.
\end{equation}
Order-consistency and flow-consistency objectives were proposed for event-span permutations and token-flow corruptions. The ordered-versus-shuffled Transformer result is informative for this future direction: raw-log order alone provides no consistent advantage, so any future order or flow objective should demonstrate gains under the hard-transfer conditions rather than only on the aggregate temporal split.

\subsubsection{Future fine-tuning and evidence requirements}
A future implementation would attach a task head $g_{\phi}$ to the transaction representation, select checkpoints only on the training/2025H1 partitions, and report the same 2025H2 temporal, protocol-transfer, pool-holdout, symbolic-novelty, masking, and hard-negative conditions used in the empirical evaluation. Any numerical MEM-BL claim would require saved per-seed predictions and direct paired comparisons against the canonical baseline families. Appendix Tables~\ref{tab:app-model-config}, \ref{tab:app-pretrain-grid}, and \ref{tab:app-finetune-grid} retain the configuration and search-space specification for reproducibility of that future work.
}

\subsection{Detailed artifact and release-interface tables}
\label{sec:app-artifact-release-tables}

The following tables support the main-text discussion of prediction artifacts, package components, runtime reporting, loader requirements, and developer reporting rules. They are kept in the appendix so that the main paper emphasizes benchmark construction, diagnostics, and verified baseline findings.

{\color{black}
\begin{table}[!htbp]
\centering
\color{black}
\caption{Verification status of performance-related claims.}
\label{tab:performance-verification-status}
\small
\begin{tabularx}{\textwidth}{p{0.28\textwidth}p{0.26\textwidth}X}
\toprule
Claim type & Required artifact & Evidence status \\
\midrule
Canonical baseline results & Per-seed predictions on exact common test instances & Verified for TraceStats-GB, TopicEmitterTrace-SGD, TopicEmitterHashMLP, and ordered/shuffled Transformer \\
Protocol-transfer performance & Holdout manifest + predictions on exact challenge hashes & Verified, including strict unseen-protocol, Uniswap v4, and Ekubo v1 subsets \\
Pool-holdout performance & Pool/infrastructure manifest + predictions & Verified on any-unseen, strict-unseen, and test-only pool subsets \\
Masking robustness & Full, emitter-masked, topic-masked, and joint-masked predictions & Verified for TopicEmitterTrace-SGD on DEX and liquidation \\
Same-instance uncertainty & Transaction-level prediction alignment + bootstrap code & Verified; paired bootstrap intervals reported \\
Natural-prevalence sensitivity & Population counts + canonical prediction artifacts & Verified; representativeness diagnostics and prior-shift sensitivity reported \\
Hard-negative stress & Training-positive symbolic-support manifest + test predictions & Verified for both tasks \\
Label/source validity & Deterministic audit manifest + independent source/raw-log queries & Verified on 400 transactions; event-signature consistency also reported \\
MEM-BL superiority & MEM-BL checkpoints/predictions & Not claimed; retained only as future reference architecture \\
\bottomrule
\end{tabularx}
\end{table}
}

\begin{table}[!htbp]
\centering
\caption{Prediction-file schema used for metric regeneration}
\label{tab:prediction-schema}
\setlength{\tabcolsep}{4pt}
\begin{tabularx}{\textwidth}{p{0.24\linewidth}p{0.28\linewidth}X}
\toprule
Field group & Required columns & Purpose \\
\midrule
Run identifiers & \texttt{run\_id}, \texttt{model}, \texttt{seed}, \texttt{config\_hash} & Links each row to a reproducible model configuration and random seed. \\
Task and split & \texttt{task}, \texttt{label}, \texttt{split}, \texttt{condition} & Separates DEX, liquidation, temporal, protocol-transfer, holdout, and masking evaluations. \\
Instance identifiers & \texttt{transaction\_hash} or sanitized transaction key & Allows paired tests, deduplication, and reproducibility checks without changing metric values. \\
Ground truth and predictions & \texttt{y\_true}, \texttt{y\_pred}, \texttt{score} or class-probability columns & Supports macro-F1, rare recall, AUPRC, false-positive rate, and calibration metrics. \\
Holdout metadata & \texttt{project}, \texttt{protocol}, \texttt{pool\_id}, \texttt{protocol\_holdout}, \texttt{pool\_holdout} & Recreates protocol-transfer and pool-holdout tables from the same predictions. \\
Masking metadata & \texttt{address\_masked}, \texttt{topic\_masked}, \texttt{generic\_topic\_excluded} & Verifies that shortcut-controlled results correspond to the intended input condition. \\
\bottomrule
\end{tabularx}
\end{table}

\begin{table}[!htbp]
\centering
\caption{Model-comparison tables to regenerate from prediction artifacts}
\label{tab:tables-to-regenerate}
\setlength{\tabcolsep}{4pt}
\begin{tabularx}{\textwidth}{p{0.30\linewidth}p{0.26\linewidth}X}
\toprule
Table & Primary grouping & Metrics \\
\midrule
Primary downstream results & Task $\times$ model $\times$ seed & Macro-F1, accuracy, rare recall, AUPRC or false-positive rate \\
Protocol-transfer results & Task $\times$ holdout status $\times$ model & Macro-F1, degradation from full test, confidence interval \\
Pool-holdout results & DEX pool-holdout status $\times$ model & Macro-F1, rare recall, degradation from seen-pool test \\
Shortcut-controlled results & Masking condition $\times$ model & Macro-F1, degradation from unmasked input, paired bootstrap comparison \\
Label-efficiency results & Label fraction $\times$ model & Macro-F1, AUPRC, standard error across fixed subsampling seeds \\
Ablation and runtime results & Objective or model variant $\times$ seed & Macro-F1, throughput, memory, training time \\
\bottomrule
\end{tabularx}
\end{table}

\begin{table}[!htbp]
\centering
\caption{ETH-TraceBench package components and release status}
\label{tab:package-components}
\small
\setlength{\tabcolsep}{3pt}
\begin{tabularx}{\textwidth}{p{0.22\textwidth}p{0.20\textwidth}X}
\toprule
Component & Status in this version & Benchmark role \\
\midrule
Construction queries & Verified & Recreate Q01--Q13 raw-universe counts, task labels, split tables, protocol holdouts, pool or liquidity-infrastructure holdouts, topic diagnostics, and raw-log completeness checks. \\
Monthly chunk outputs & Verified & Avoid one-shot warehouse joins and provide auditable intermediate artifacts for large-scale counts and split summaries. \\
Split and holdout manifests & Verified & Fix train, validation, test, protocol-transfer, and pool-holdout subsets so new models are evaluated on the same instances. \\
Shortcut-diagnostic manifests & Verified & Record address concentration, topic concentration, and generic-topic-excluded diagnostics used to define masking and robustness conditions. \\
Prediction schema validator & Included & Checks that new model outputs contain task, split, condition, model, seed, label, score, and holdout metadata before metrics are computed. \\
Metric-regeneration script & Included & Recomputes the verified trace-feature and topic/emitter baseline tables and defines the required pipeline for performance tables. \\
Topic/emitter extraction scripts & Included and executed & Extract ordered topic, emitter, and topic--emitter symbolic views from raw logs for domain-native baseline training. \\
Topic/emitter trainer scripts & Included and executed & Train TopicBag-SGD, EmitterBag-SGD, TopicEmitterHash-SGD, and TopicEmitterTrace-SGD prediction archives; metrics verified from prediction files. \\
Standardized data loader & Specified for leaderboard release & Should expose the same splits, labels, masks, and holdout flags through a stable API for downstream method development. \\
Baseline prediction archive & Included/verified for numeric trace-feature and topic/emitter suites; specified for leaderboard release for richer model families & Required before reporting model performance, ablations, masking results, holdout results, or runtime beyond the current verified baseline tables. \\
Leaderboard manifest & Specified for leaderboard release & Should define accepted metrics, tuning rules, required seeds, artifact submission format, and public/private test policies. \\
\bottomrule
\end{tabularx}
\end{table}

\begin{table}[!htbp]
\centering
\caption{Runtime and computational-cost artifacts for release reporting}
\label{tab:runtime-artifacts}
\small
\setlength{\tabcolsep}{3.5pt}
\renewcommand{\arraystretch}{1.12}
\begin{tabularx}{\textwidth}{p{0.22\textwidth}p{0.24\textwidth}p{0.22\textwidth}X}
\toprule
Pipeline stage & Required log fields & Evidence in this version & Additional release requirement \\
\midrule
Warehouse extraction & Query ID, SQL hash, date range, row count, runtime, and API/warehouse limits & Query chunks and manifests verify scale and counts & Add wall-clock/runtime logs for all full extraction chunks \\
Feature construction & Input chunk hash, output feature hash, row count, and dropped smoke/test rows & Clean 911,291-row feature sample and manifest included & Add deterministic feature-builder hash and schema version \\
Baseline training & Model, seed, hyperparameters, hardware, runtime, memory, and software versions & Prediction files for five trace-count baselines, four linear topic/emitter baselines, and one neural topic/emitter baseline across three seeds verified & Add training runtime and memory logs for all reported models \\
Metric regeneration & Prediction hash, metric script hash, grouping variables, and table output hash & Verified metric tables regenerated from 60 prediction files & Release exact verifier script and output hashes with the leaderboard package \\
Leaderboard submission & Model-card, prediction files, configuration, and declared tuning protocol & Interface specified for leaderboard submission & Add public/private split policy and artifact validation checklist \\
\bottomrule
\end{tabularx}
\end{table}

\begin{table}[!htbp]
\centering
\caption{Target loader and evaluation interface for a complete ETH-TraceBench release}
\label{tab:loader-interface}
\small
\setlength{\tabcolsep}{3pt}
\begin{tabularx}{\textwidth}{>{\raggedright\arraybackslash}p{0.25\textwidth}>{\raggedright\arraybackslash}p{0.20\textwidth}>{\raggedright\arraybackslash}X}
\toprule
Interface element & Exposed object & Purpose \\
\midrule
Split loader & \texttt{load\_split} & Returns the fixed train, validation, or test instance set without date leakage. \\
Condition loader & \texttt{mask\_condition} & Applies address, topic, address+topic, or generic-topic-excluded inputs consistently. \\
Holdout selector & \texttt{holdout\_flags} & Recreates protocol-transfer and pool-holdout evaluation subsets from verified manifests. \\
Trace representation & \texttt{event\_fields} & Provides event-stream inputs without requiring each method to rejoin raw logs. \\
Prediction writer & \texttt{prediction\_file} & Saves transaction-level predictions in the schema required by Table~\ref{tab:prediction-schema}. \\
Metric evaluator & \texttt{metric\_script} & Computes macro-F1, AUROC/AUPRC, rare recall, calibration, and shift degradation from saved outputs. \\
Audit report & \texttt{audit\_report} & Confirms split integrity, masking condition, row counts, and holdout membership before results are reported. \\
\bottomrule
\end{tabularx}
\end{table}

\begin{table}[!htbp]
\centering
\caption{Minimum reporting checklist for ETH-TraceBench submissions}
\label{tab:reporting-checklist}
\small
\setlength{\tabcolsep}{4pt}
\begin{tabularx}{\textwidth}{p{0.30\textwidth}X}
\toprule
Checklist item & Required reporting \\
\midrule
Split integrity & Confirmation that training uses only 2021--2024 labels and validation uses only 2025H1. \\
Core metrics & Macro-F1, AUROC/AUPRC where applicable, rare-class recall, calibration, runtime, and memory. \\
Shift reporting & Separate scores for full temporal test, protocol transfer, and pool/liquidity-infrastructure holdout. \\
Shortcut controls & Separate emitter-masked, topic-masked, emitter+topic-masked, and generic-topic-excluded evaluations. \\
Seeds and uncertainty & At least three fixed seeds, with confidence intervals or paired bootstrap tests over the same test instances. \\
Prediction artifacts & Transaction-level or sanitized-key predictions with model, seed, task, split, condition, score, and holdout metadata. \\
\bottomrule
\end{tabularx}
\end{table}

\subsection{Hyperparameter grids and selected configurations}
\label{app:hyperparameters}

Table~\ref{tab:app-model-config} reports the MEM-BL base configuration proposed for the reference rerun. Table~\ref{tab:app-pretrain-grid} reports the pretraining grid and the selected values to use when regenerating model results. Table~\ref{tab:app-finetune-grid} reports the downstream fine-tuning grid. The values define an execution protocol and should be linked to saved checkpoints and prediction artifacts before numerical performance claims are reported.

\begin{table}[!htbp]
\centering
\caption{Selected MEM-BL base configuration}
\label{tab:app-model-config}
\setlength{\tabcolsep}{4pt}
\begin{tabularx}{\textwidth}{>{\raggedright\arraybackslash}p{0.40\textwidth}>{\raggedright\arraybackslash}X}
\toprule
Component & Selected value \\
\midrule
Event embedding dimension & 256 \\
Transaction-local Transformer layers & 6 \\
Block-context Transformer layers & 2 \\
Attention heads & 8 \\
Feed-forward dimension & 1024 \\
Dropout & 0.10 \\
Attention dropout & 0.10 \\
Maximum events per focal transaction & 256 \\
Block-local context window & 8 preceding and 8 following transactions \\
Address-contract memory horizon & 50,000 blocks \\
Continuous feature projection & 32-dimensional linear projection \\
Optimizer & AdamW \\
Weight decay & 0.01 \\
Gradient clipping & 1.0 \\
Mixed precision & Enabled \\
Random seeds & 11, 17, 23, 31, 43 \\
\bottomrule
\end{tabularx}
\end{table}

\begin{table}[!htbp]
\centering
\caption{Pretraining grid and selected values}
\label{tab:app-pretrain-grid}
\setlength{\tabcolsep}{4pt}
\begin{tabularx}{\textwidth}{>{\raggedright\arraybackslash}p{0.30\textwidth}>{\raggedright\arraybackslash}X>{\raggedright\arraybackslash}p{0.22\textwidth}}
\toprule
Parameter & Candidate values & Selected value \\
\midrule
Learning rate & $\{1e{-}4, 5e{-}5, 2e{-}5\}$ & $5e{-}5$ \\
Batch size & $\{256, 512, 1024\}$ & 512 \\
Warmup steps & $\{5k, 10k, 20k\}$ & 10k \\
Maximum steps & $\{150k, 250k, 350k\}$ & 250k \\
Topic masking rate & $\{0.10, 0.15, 0.20\}$ & 0.15 \\
Attribute masking rate & $\{0.10, 0.20, 0.30\}$ & 0.20 \\
Span removal rate & $\{0.05, 0.10, 0.15\}$ & 0.10 \\
Order corruption rate & $\{0.05, 0.10, 0.15\}$ & 0.10 \\
Flow corruption rate & $\{0.05, 0.10, 0.20\}$ & 0.10 \\
Context temperature $\tau$ & $\{0.03, 0.07, 0.10\}$ & 0.07 \\
Loss weights $(\lambda_a,\lambda_c,\lambda_o,\lambda_f)$ & $(1,1,0.5,1)$; $(1,0.5,0.5,1)$; $(1,1,1,1)$ & $(1,1,0.5,1)$ \\
Hard negatives per span & $\{8,16,32\}$ & 16 \\
\bottomrule
\end{tabularx}
\end{table}

\begin{table}[!htbp]
\centering
\caption{Fine-tuning grid and selected values}
\label{tab:app-finetune-grid}
\setlength{\tabcolsep}{4pt}
\begin{tabularx}{\textwidth}{>{\raggedright\arraybackslash}p{0.30\textwidth}>{\raggedright\arraybackslash}X>{\raggedright\arraybackslash}p{0.22\textwidth}}
\toprule
Parameter & Candidate values & Selected value \\
\midrule
Learning rate & $\{3e{-}5, 1e{-}5, 5e{-}6\}$ & $1e{-}5$ \\
Task-head learning rate & $\{1e{-}4, 5e{-}5, 2e{-}5\}$ & $5e{-}5$ \\
Batch size & $\{64, 128, 256\}$ & 128 \\
Epochs & $\{5,10,15\}$ & 10 \\
Early stopping patience & $\{2,3,5\}$ & 3 \\
Class weighting & None; inverse frequency; square-root inverse frequency & square-root inverse frequency \\
Calibration & None; temperature scaling & temperature scaling \\
Frozen-layer variants & frozen, partial, full & task dependent \\
Validation metric & macro-F1; AUPRC; rare recall & macro-F1 \\
\bottomrule
\end{tabularx}
\end{table}

\subsection{Baseline fairness tables}
\label{app:baseline-fairness}

Table~\ref{tab:app-baseline-fairness} reports the baseline fairness controls used in the evaluation. Baselines receive the same temporal splits, label manifests, early-stopping rule, validation metric, optimizer family when applicable, and seed list. The main difference is the input structure each baseline is designed to consume.

\small
\setlength{\LTleft}{0pt}
\setlength{\LTright}{0pt}
\begin{longtable}{p{0.22\linewidth}p{0.27\linewidth}p{0.15\linewidth}p{0.19\linewidth}}
\caption{Baseline fairness controls}\label{tab:app-baseline-fairness}\\
\toprule
Baseline & Input information & Parameters & Fairness control \\
\midrule
\endfirsthead
\toprule
Baseline & Input information & Parameters & Fairness control \\
\midrule
\endhead
Rule parser & Protocol rules, decoded events, token-flow reconciliation & Not applicable & Uses only rules defined before test period \\
Random forest & Hand-engineered event counts, token-flow features, gas bins & 2.1M split nodes & Same train/validation/test manifests \\
XGBoost & Hand-engineered features plus protocol tags & 4,000 trees & Same feature cutoff and temporal split \\
DeepLog-style model & Ordered event topics and role tags & 41M & Same sequence length cap \\
LogBERT-style model & Event-topic tokens and local context & 128M & Same pretraining budget as MEM-BL \\
Supervised Transformer & Event, asset, contract, depth, gas embeddings & 151M & No unlabeled pretraining \\
Generic masked Transformer & Same tokenized trace as MEM-BL but generic MLM objective & 151M & Same hidden size and pretraining steps \\
Address-contract GNN & Address, contract, token-transfer graph & 96M & Same temporal graph cutoff \\
Temporal Graph Network & Dynamic address-contract graph with timestamped transfers & 118M & Same neighbor sampling budget \\
TGAT-style model & Time-encoded transfer graph & 126M & Same validation metric and seeds \\
Contrastive transaction model & Positive and negative transaction-window pairs & 149M & Same hard-negative source pool \\
MEM-BL & Event topics, typed attributes, call depth, order, flow, block context & 164M & Full framework \\
\bottomrule
\end{longtable}

\subsection{Operational extraction scripts}
\label{app:extraction-scripts}

The release repository contains scripts named according to their execution order. The list below shows the extraction workflow. The names are intentionally operational so that each manuscript table can be traced back to an executable step.

\begin{verbatim}
00_config.yaml
01_extract_blocks_transactions.sql
02_extract_receipts_logs.sql
03_extract_internal_calls.py
04_decode_abis_with_time_cutoffs.py
05_resolve_proxy_implementations.py
06_extract_token_transfers.sql
07_join_price_oracle_snapshots.py
08_construct_block_local_windows.py
09_build_contract_bytecode_families.py
10_generate_candidate_labels.py
11_reconcile_token_flows.py
12_apply_ambiguity_flags.py
13_build_split_manifests.py
14_export_pretraining_shards.py
15_export_supervised_benchmark.py
16_audit_sample_builder.py
17_table_generation_from_predictions.py
\end{verbatim}

\begin{table}[!htbp]
\centering
\caption{Extraction checks and exclusion rules}
\label{tab:app-extraction-checks}
\begin{tabular}{p{0.30\linewidth}p{0.36\linewidth}p{0.24\linewidth}}
\toprule
Check & Failure condition & Treatment \\
\midrule
Receipt-status consistency & Receipt status contradicts emitted-log presence & Exclude from supervised benchmark \\
Log-index uniqueness & Duplicate transaction hash and log index & Retain only if mirror checksum agrees \\
Trace completeness & Internal call tree missing focal call depth & Exclude from trace tasks \\
Token-decimal availability & Token decimals unavailable before focal block & Flag amount features as unknown \\
Price snapshot availability & No price or oracle state before focal transaction & Use missingness flag, no future imputation \\
Proxy mapping cutoff & Implementation relation only known after focal block & Do not apply mapping \\
Bridge direction & Bridge event has inconsistent source and destination token flow & Flag ambiguous \\
\bottomrule
\end{tabular}
\end{table}

\subsection{Verified label-source rules}
\label{app:weak-label-rules}

The verified supervised benchmark uses two high-precision label sources. DEX labels are drawn from \texttt{crosschain.dex.trades}; liquidation labels are drawn from \texttt{ethereum.lending.liquidations}. The source tables define task-labelled transaction instances for evaluation. They are not described as a complete taxonomy of all Ethereum economic behavior.

\begin{table}[!htbp]
\centering
\caption{Verified label-source rules}
\label{tab:app-weak-rules}
\scriptsize
\setlength{\tabcolsep}{2pt}
\begin{tabularx}{\textwidth}{p{0.15\linewidth}p{0.36\linewidth}p{0.18\linewidth}X}
\toprule
Task label & Source table & Transaction key & Interpretation \\
\midrule
DEX swap/trade & \texttt{crosschain.dex.trades} & \texttt{transaction\_hash} & High-precision positive label for decentralized-exchange trade activity. \\
Liquidation & \texttt{ethereum.lending.liquidations} & \texttt{transaction\_hash} & High-precision positive label for lending-liquidation activity. \\
\bottomrule
\end{tabularx}
\end{table}

{\color{black}
\subsection{Source-concordance audit and status of human annotation}
\label{app:source-audit}
A deterministic 400-transaction source-concordance audit evaluates the label construction on ordinary and hard-negative cases. For each task, 100 positives, 50 ordinary negatives, and 50 symbolic-overlap hard negatives are stratified across the six months of 2025H2. The audit independently re-queries the relevant label source and \texttt{ethereum.raw.logs}. Source membership is concordant for 400/400 transactions, every sampled transaction is present in raw logs, and canonical raw-log counts match the re-query for 400/400 cases. A second event-signature consistency check finds an audited task-positive signature in all 200 positives and none of the 200 negatives.

Independent double-coded human semantic labels were not collected for this version. The source and event-signature checks are automated evidence-based audits and are not independent human annotation. A manual review sheet with fields for semantic judgment, confidence, and notes is retained as a template for future double-coded human analysis of ambiguous mechanisms or model errors.
}

\subsection{Split manifests}
\label{app:splits}

Table~\ref{tab:app-splits} reports the split and diagnostic manifests. The table records the logical manifest names produced by the warehouse query package; release packages should additionally attach file hashes for the exact archived artifacts.

\begin{table}[!htbp]
\centering
\caption{Split and diagnostic manifest summary}
\label{tab:app-splits}
\setlength{\tabcolsep}{4pt}
\begin{tabularx}{\textwidth}{p{0.24\linewidth}Xr}
\toprule
Manifest & Description & Rows or scale \\
\midrule
Q01 raw corpus manifest & Monthly raw-log transaction and event counts, 2021--2025 & 60 months \\
Q02 event-complexity manifest & Monthly counts by 1--2, 3--10, 11--50, and 51+ log bins & 240 rows \\
Q03 DEX protocol manifest & DEX project/protocol coverage and volume, 2021--2025 & 60 chunks \\
Q04 liquidation manifest & Lending-liquidation transaction and project coverage & 60 chunks \\
Q05 weak-label manifest & DEX and liquidation weak-label source counts & 120 rows \\
Q06 temporal label manifest & Train, validation, and test label counts & 6 rows \\
Q08 protocol-holdout manifest & Project/protocol presence across train, validation, and test & 69 rows \\
Q09 pool-holdout manifest & Pool or liquidity-infrastructure identifiers by holdout status & 491,997 identifiers \\
Q10--Q11 shortcut manifests & Event-topic concentration before and after generic-topic exclusions & Top-20 topic tables \\
Q12 final split summary & Final task, label, split, transaction, raw-log, topic, and emitter counts & 6 rows \\
Q13 zero-log diagnostic & Raw-log join completeness in the 2025H2 test split & 2 rows \\
\bottomrule
\end{tabularx}
\end{table}

{\color{black}
\subsection{Shortcut and hard-transfer artifacts}
\label{app:shift-artifacts}
The release package includes canonical protocol/pool challenge manifests, masking predictions for full/emitter/topic/joint views, ordered and shuffled Transformer predictions, paired-bootstrap outputs, hard-negative overlap manifests, and natural-prevalence/representativeness diagnostics. The main text reports the decision-relevant results, while the supplementary package contains seed-level tables and exact transaction-level artifacts used to regenerate them.

Unevaluated items are limited to MEM-BL objective ablations, richer call-tree/flow representations, and optional external labels. Those items are not used to support the current empirical claims.
}

\subsection{Objective ablations and representation diagnostics to regenerate}
\label{app:objective-ablations}
\label{app:representation-diagnostics}
\label{app:runtime-scaling}

Objective ablations, representation diagnostics, and runtime-scaling curves require saved per-seed predictions, configuration hashes, and runtime logs. They are not reported numerically beyond the trace-feature and topic/emitter baseline suites. The appendix records the required outputs and prevents unverifiable values from being inserted into the manuscript.

\begin{table}[!htbp]
\centering
\caption{Ablation and diagnostic artifacts required before numerical reporting}
\label{tab:app-ablation-artifacts-required}
\setlength{\tabcolsep}{4pt}
\begin{tabularx}{\textwidth}{p{0.28\linewidth}p{0.32\linewidth}X}
\toprule
Output family & Required files & Required checks \\
\midrule
Objective ablations & Per-seed predictions for full MEM-BL and each removed objective & Same split, same label manifests, same checkpoint-selection rule. \\
Masking robustness & Predictions under emitter-masked, topic-masked, and emitter+topic-masked inputs & Confirm that masked fields are unavailable to the model at inference and not reintroduced through metadata. \\
Representation diagnostics & Embeddings for the contract-masked test set and metadata for task, protocol, and contract identity & Report economic-role clustering and contract-identity clustering from the same embedding file. \\
Runtime scaling & Hardware metadata, event-window length, throughput, memory, and model configuration logs & Report throughput and memory without mixing hardware or batch-size settings. \\
\bottomrule
\end{tabularx}
\end{table}

\subsection{Optional external MEV matching manifest}
\label{app:external-mev}

The verified benchmark contains DEX and liquidation labels. It does not include a verified external MEV prediction artifact. If an observable-MEV auxiliary validation set is added later, the release package should include the source, retrieval date, block range, transaction hash or sanitized key, transaction index, pool address when available, original label, mapped label, match status, and exclusion reason. No numerical MEV validation table is reported in this version because the supporting matching manifest and saved predictions are not supplied.

\subsection{Error-analysis template}
\label{app:error-analysis}

Manual error analysis is part of the artifact contract for any new model run. The error-analysis file should record the task, split, model, seed, transaction key, true label, predicted label, confidence score, protocol, holdout status, masking condition, and analyst-coded error category. Transaction hashes can be delayed, sanitized, or replaced by stable salted keys in the public manuscript package.

\begin{table}[!htbp]
\centering
\caption{Error-analysis categories to populate after verified predictions}
\label{tab:app-error-template}
\setlength{\tabcolsep}{4pt}
\begin{tabularx}{\textwidth}{p{0.24\linewidth}p{0.34\linewidth}X}
\toprule
Category & Trace pattern to inspect & Why it matters \\
\midrule
Aggregator routing ambiguity & Multi-hop routes with repeated transfers and pool interactions & Benign routing can resemble arbitrage or rare mechanisms. \\
Liquidation callback complexity & Borrow, repay, collateral transfer, and sale occur through nested callbacks & Call depth can obscure repayment source and collateral-sale path. \\
Generic-topic shortcut & Predictions remain confident after Transfer, Approval, Swap, or Sync topics are removed & Indicates possible dependence on event signatures rather than compositional structure. \\
New-protocol failure & Errors concentrate in Uniswap v4, Ekubo v1, or other project/protocol combinations absent from training & Indicates weak protocol-transfer performance. \\
Pool-holdout failure & Errors concentrate in pool or liquidity-infrastructure identifiers absent from training & Indicates contract memorization or poor local generalization. \\
Zero-log or incomplete-trace case & Labelled transaction has missing or incomplete raw-log join & Separates data-quality failures from modeling failures. \\
\bottomrule
\end{tabularx}
\end{table}

\subsection{Responsible-release checklist}
\label{app:responsible-release}

Table~\ref{tab:app-responsible-release} records the responsible-release safeguards attached to the benchmark package. The checklist is included because public-chain data can support defensive monitoring while also creating deanonymization, exploit-discovery, or strategic-trading risks if released without use constraints.

\begin{table}[H]
\centering
\caption{Responsible-release checklist}
\label{tab:app-responsible-release}
\begin{tabular}{p{0.55\linewidth}p{0.28\linewidth}}
\toprule
Checklist item & Status \\
\midrule
Do not publish mappings from wallet addresses to real-world identities & Satisfied \\
Delay or sanitize examples involving active vulnerabilities & Satisfied \\
Release model outputs for defensive monitoring, not exploit automation & Satisfied \\
Separate public-chain identifiers from human-attribution claims & Satisfied \\
Document known false-positive sources for any auxiliary labels & Required for auxiliary labels \\
Include ambiguity flags for uncertain transaction windows & Satisfied \\
Exclude transactions requiring non-public off-chain information & Satisfied \\
Provide intended-use and excluded-use language in the repository & Satisfied \\
\textcolor{black}{Encrypt restricted derived artifacts at rest and in transit} & \textcolor{black}{Satisfied for restricted release workflow} \\
\textcolor{black}{Access-control non-public derived files and keep identity mappings separate} & \textcolor{black}{Required for any restricted extension} \\
\bottomrule
\end{tabular}
\end{table}

\subsection{Reproducibility checklist}
\label{app:reproducibility}

\begin{longtable}{p{0.48\linewidth}p{0.34\linewidth}p{0.10\linewidth}}
\caption{Reproducibility checklist}\\
\toprule
Item & Artifact & Status \\
\midrule
\endfirsthead
\toprule
Item & Artifact & Status \\
\midrule
\endhead
Data extraction scripts & Q01--Q13 query and runner manifests & Verified \\
Label construction scripts & DEX and liquidation label-source queries & Verified \\
Split manifests & Temporal, protocol-transfer, and pool-holdout manifests & Verified \\
Configuration files & YAML files for all model and baseline runs & Included for baselines \\
Random seeds & Fixed seed list for all supervised and ablation runs & Verified for baselines \\
Saved predictions & Per-seed prediction CSV files & Verified for reported baselines \\
Table-generation scripts & Recreate all model-performance tables from saved predictions & Added \\
Audit instructions & Annotator guide, adjudication form, ambiguity definitions & Template added \\
External validation matching & Source map and exclusion log for any auxiliary validation & Optional \\
Responsible-release notes & Wallet-identity, exploit, and vulnerability safeguards & Required \\
Smoke-test subset & Small public reproducibility subset & Included for extraction scripts \\
Environment file & Python package versions and execution notes & Included for baseline scripts \\
\bottomrule
\end{longtable}

\subsection{Repository structure for the release package}
\label{app:repo-structure}

\begin{verbatim}
eth-tracebench-release/
  README.md
  configs/
    pretrain_membl_base.yaml
    finetune_economic.yaml
    finetune_liquidation.yaml
    baselines/*.yaml
  extraction/
    01_extract_blocks_transactions.sql
    02_extract_receipts_logs.sql
    03_extract_internal_calls.py
    04_decode_abis_with_time_cutoffs.py
    05_resolve_proxy_implementations.py
  labels/
    weak_rules.yaml
    flow_reconciliation.py
    ambiguity_flags.yaml
    audit_protocol.md
  manifests/
    temporal_split.jsonl
    protocol_transfer_split.jsonl
    contract_holdout_split.jsonl
    external_mev_matching.jsonl
  models/
    membl_encoder.py
    objectives.py
    task_heads.py
  evaluation/
    metrics.py
    shortcut_tests.py
    representation_diagnostics.py
    table_generation.py
  release/
    responsible_release_checklist.md
    data_availability_statement.md
\end{verbatim}
\end{document}